\documentclass{article}
\usepackage{arxiv}
\usepackage{hyperref}
\usepackage{url}            % simple URL typesetting
\usepackage{booktabs}  
\usepackage{amsmath}% http://ctan.org/pkg/amsmath
\usepackage{kbordermatrix}
\usepackage{subfigure}
\usepackage{graphicx}
\usepackage[square,sort,comma,numbers]{natbib}
\usepackage{dsfont}

\title{Enhancing Symbolic Regression and Universal Physics-Informed Neural Networks with Dimensional Analysis}
\author{Lena Podina \\
    Cheriton School of Computer Science\\
    University of Waterloo\\
    200 University Avenue West\\
    Waterloo, Ontario, N2L 3G1\\
	\And
    Diba Darooneh \\
    Department of Electrical and Computer Engineering \\
    University of Waterloo\\
    200 University Avenue West\\
    Waterloo, Ontario, N2L 3G1\\
	\And
	Joshveer Grewal \\
    Electrical Engineering and Computer Science Department\\
    University of Michigan\\
    1109 Geddes Avenue\\
    Ann Arbor, MI, United States
	\And
	Mohammad Kohandel \\
    Department of Applied Mathematics\\
    University of Waterloo\\
    200 University Avenue West\\
    Waterloo, Ontario, N2L 3G1\\
}

\date{}

\renewcommand{\headeright}{}
\renewcommand{\undertitle}{}
\renewcommand{\shorttitle}{Enhancing Symbolic Regression and UPINNs with Dimensional Analysis}

\begin{document}
\maketitle

\begin{abstract}
    
In engineering and applied mathematics, developing accurate mathematical models to predict and understand real-world phenomena is of utmost importance. Symbolic regression is a useful machine learning-based tool to fit models but it can be computationally expensive. We present a new method for enhancing symbolic regression for differential equations via dimensional analysis, specifically the Buckingham $\Pi$ theorem and Ipsen's method. Since symbolic regression often suffers from high computational costs and overfitting, nondimensionalizing datasets reduces the number of input variables, simplifies the search space, and ensures that derived equations are physically meaningful. As a first step, we combine dimensional analysis with the PySR symbolic regression algorithm to show that dimensional analysis improves the accuracy of recovering algebraic equations. The results demonstrate that transforming data into a dimensionless form significantly improves the training and test error of the symbolic expressions found. Then, as our main contribution, we perform nondimensionalization guided by Ipsen's method. We then incorporate the nondimensionalized equation into a pipeline combining Universal Physics-Informed Neural Networks and symbolic regression to recover the unknown term when a differential equation is only partially known. We find that symbolic regression is able to better recover the unknown term after nondimensionalizing the data, under both noisy and noiseless conditions. These findings suggest that integrating dimensional analysis with symbolic regression can significantly lower computational costs and increase accuracy, providing a robust framework for automated discovery of governing equations in complex systems when data is limited.
\end{abstract}

\section{Introduction}

In many engineering disciplines, such as aerodynamics, heat transfer, and materials science, developing accurate, interpretable mathematical models is crucial for control and optimization. Engineers often rely on a combination of empirical data and physical laws to build such models~\cite{willard2022integrating,bergen2019machine,hsieh2009machine}. However, deriving governing equations from raw experimental data can be difficult, especially when the underlying relationships are nonlinear or partially unknown. Symbolic regression has gained popularity as a method for discovering mathematical models directly from data~\cite{udrescu2020ai,valipour2021symbolicgpt,tenachi2023deep,keren2023computational}. Unlike traditional regression, which assumes a specific model form, symbolic regression searches for the best-fitting closed-form parsimonious equation for the dataset without predefined assumptions. Symbolic regression also has the advantage of being more interpretable than model-fitting via neural networks~\cite{fan2021interpretability}. The flexibility of symbolic regression algorithms makes them powerful for uncovering hidden relationships in complex datasets. However, the method's exhaustive search over possible equations can be computationally expensive, leading to slow performance and the risk of overfitting. Symbolic regression is generally an NP-hard problem~\cite{virgolin2022symbolic}.

Dimensional analysis, a widely used technique in mathematics and engineering~\cite{ferro2019assessing}, simplifies algebraic or differential equations by reducing the number of variables involved. This allows the true underlying equations and hidden parameters to be recovered with higher accuracy and with less computational overhead. It also ensures that the recovered equation adheres to the physical laws. Dimensional analysis can be performed using different methods e.g. manual nondimensionalization, Buckingham $\Pi$ theorem~\cite{sonin2001physical}, and Ipsen's method~\cite{clark2022alternate}. Manual nondimensionalization involves scaling each dimensional variable by a hand-chosen value to remove its units, then substituting the non-dimensional variables for the dimensional ones. This removes the units but may not reduce the number of free parameters. The Buckingham $\Pi$ theorem introduces an automated procedure to minimize the possible number of dimensionless constants ($\pi$ terms) that the equation can be expressed with. Although there are many choices of sets of $\pi$ terms, it is difficult to directly control the final equation, although it will have the minimal number of free variables. Ipsen's method is very similar to the Buckingham $\Pi$ theorem, but enables straightforward user selection of the dimensionless constants. This method is useful when one is interested in reducing (or maximizing) the number of free variables that appear in a specific term in the equation. Additionally, within mathematical modelling for engineering applications, it is common to reduce a problem to a similar but simpler formulation~\cite{schmidt2009distilling,muagurean2023three,ferro2019assessing}.

Symbolic regression algorithms have been built using various approaches such as genetic programming, sparse regression, neural networks and language models~\cite{valipour2021symbolicgpt,kamienny2022end,augusto2000symbolic,cranmerInterpretableMachineLearning2023}, with some works~\cite{udrescu2020ai,tenachi2023deep,keren2023computational} integrating dimensional analysis into their workflow. However, none of these approaches integrate dimensional analysis in a partial physical knowledge setting, and systematically show the effect of nondimensionalization on a symbolic regression algorithm such as PySR~\cite{cranmerInterpretableMachineLearning2023}. As a preliminary proof of concept, we test that dimensional analysis in fact improves the performance of symbolic regression. Then, we propose a novel method for increasing efficiency and accuracy of symbolic regression under partial-knowledge by first nondimensionalizing the data. We stay in a regime where some terms in a differential equation are known, and some need to be learned, which is common in many engineering and mathematical modelling applications. Ipsen's method of nondimensionalization allows for the fine control of the variables that appear in different components of the differential equation. Then, we apply Universal Physics-Informed Neural Networks (UPINNs)~\cite{pmlr-v202-podina23a} to learn individual unknown terms in an equation. This approach allows the user to reduce the number of input variables to the unknown components while ensuring the dimensional consistency of expressions. Due to nondimensionalization, our strategy also potentially avoids calculations involving high-order or low-order numbers. The true underlying expression can be found by undoing the nondimensionalization.

By integrating dimensional analysis with any symbolic regression workflow, we can significantly enhance the applicability of data-driven discovery in engineering. The symbolic search space becomes significantly smaller because candidate expressions only need to be constructed from dimensionless groups rather than the original variables, and dimensionally inconsistent expressions are automatically eliminated from consideration. In turn, the reduced search space decreases the risk of overfitting and spurious symbolic relationships. Our contributions in this paper are:

\begin{itemize}
    \item Demonstration of how the Buckingham $\Pi$ theorem can be combined with any symbolic regression algorithm. Specifically, when paired with PySR~\cite{cranmerInterpretableMachineLearning2023} symbolic regression, this combination recovers the correct expressions with greater accuracy.
    \item Introduction of a novel pipeline that integrates dimensional analysis with Universal Physics-Informed Neural Networks (UPINNs)~\cite{pmlr-v202-podina23a} and symbolic regression. This pipeline can be used in many fields of mathematical modelling where some physics is known and some needs to be learned, such as engineering, materials science, and computational biology. We showcase and validate the performance of this pipeline in three ordinary differential equations, under both noisy and noiseless data. 

\end{itemize}

We note that unlike statistical compression methods such as PCA or latent-variable embeddings, dimensional analysis preserves dimensional consistency, physical interpretability, and invariant relationships between quantities. In our framework, the objective is not to simply reduce the dimensionality of the dynamical system, but rather to reduce the number of independent physically admissible inputs entering the unknown term. This constrains the symbolic regression search space in a physically meaningful manner while allowing the recovered expressions to be mapped back to the original governing equations. 

% Our main contribution is the  This approach achieves a minimal number of variables in the unknown terms upon nondimensionalization, thereby enhancing the efficiency and effectiveness of both UPINNs and symbolic regression. Additionally, 

% \section{Related Work}

\section{Background}
\label{sec:background_nondim}

In this section, we describe the problem of symbolic regression and previous work in the area (Section~\ref{sec:symreg}), the Buckingham $\Pi$ method of nondimensionalization (Section~\ref{sec:buckingham}), and Ipsen's method of nondimensionalization (Section~\ref{sec:ipsen}), including how it differs from the Buckingham $\Pi$ method and its advantages. Finally, we provide a brief overview of UPINNs in Section~\ref{sec:upinns}.

\subsection{Symbolic Regression}\label{sec:symreg}

Symbolic regression is a powerful technique for discovering mathematical equations and differential equations (DEs) directly from data, integrating various methods to optimize search efficiency and ensure physical coherence of the derived models. Large classifications of symbolic regression techniques include: genetic programming; neural network and regression approaches; large language models; reinforcement learning.

Traditional approaches to symbolic regression include genetic programming, brute force~\cite{korns2013baseline}, Monte Carlo tree search~\cite{sun2022symbolic}, and optimization with hand-crafted features~\cite{brunton2016discovering}. In genetic programming, equations are considered strings of information much like a gene, and new equations for a particular dataset are generated by ``crossing over'' the best-fit equations. Notable works that use this method are~\cite{augusto2000symbolic,mckay1995using,Quade_2016}. However, this method suffers from several drawbacks. Firstly, it is fairly slow compared to other approaches~\cite{kamienny2022end}. Secondly, it is difficult to incorporate prior knowledge into the generation. 
The EPDE framework~\cite{maslyaev2021partial} is an evolutionary algorithm that operates without a predefined library of functions, utilizing basic terms to discover both differential and algebraic equations. It employs a multi-objective optimization approach to select the most effective equations and includes a custom solver that handles noisy data and supports the visualization of the discovery process. SINDy~\cite{brunton2016discovering} identifies dynamical systems by finding sparse representations of governing equations from data. When integrated with dimensional analysis, SINDy ensures that the equations are dimensionally consistent, which is crucial for maintaining physical accuracy in complex systems.

Approaches based on neural networks, such as reinforcement learning and language models, have also been widely used. Large language model-based approaches include works such as~\cite{valipour2021symbolicgpt, kamienny2022end}. SymbolicGPT uses a decoder-only approach to generate a skeleton first, and then populate constants with real values. The work by~\cite{kamienny2022end} follows a similar approach, but generates equations with constants embedded, skipping the skeleton generation. Although language-model-based generation of equations for symbolic regression has been successful, it can be difficult to directly control or constrain these methods to generate expressions with consistent units, or enforce other constraints.
Finally, reinforcement learning based approaches have been widely used by works such as~\cite{tenachi2023deep,li2024neuralguideddynamicsymbolicnetwork,petersen2021deep} due to their flexibility in choosing the reward function and ability to enforce both hard constraints on the search space, and soft constraints that are formulated via the reward function. Petersen et al.~\cite{petersen2021deep} train a recurrent neural network which creates the expression tree, and outperforms Eureqa~\cite{schmidt2009distilling}, one of the gold-standard methods that use this approach. 

A number of works integrate dimensional analysis or nondimensionalization within the symbolic regression algorithm~\cite{bakarji2022dimensionally,udrescu2020ai,tenachi2023deep,keren2023computational}. This has been done mainly to ensure that the output of the symbolic regression is physically consistent, and to reduce the number of independent variables in the problem. However, this integration in past works has been limited and is not always tailored to the problem at hand. AI Feynman~\cite{udrescu2020ai} does incorporate dimensional analysis, but the dimensions for variables are fixed and analyzed automatically, not allowing user input or partial knowledge. $\Phi$-SO~\cite{tenachi2023deep} incorporates dimensional analysis but does not nondimensionalize datasets before fitting them, which can be beneficial for both resetting the scale of the data, and reducing the number of variables. SciMED~\cite{keren2023computational} integrates dimensional analysis with symbolic regression and applies it to both algebraic and differential equations, but its application is restricted to one-side differential equations. No symbolic regression method, to our knowledge, has incorporated nondimensionalization in a partial-knowledge scenario using Universal Differential Equations~\cite{rackauckas2020universal} or Universal Physics-Informed Neural Networks~\cite{pmlr-v202-podina23a}.

The integration of PINNs with dimensional analysis has been explored to some extent in previous works~\cite{qi2025ndawl,yang2024data} but has not been applied in a partial knowledge setting such as UPINNs~\cite{pmlr-v202-podina23a}.
NDAWL-PINNs use nondimensionalization primarily in the loss function~\cite{qi2025ndawl}. The work by~\cite{yang2024data} applies dimensional analysis to a physics problem but does not enforce the differential equation constraints via a PINNs style loss function. A recent work~\cite{chandra2024role} shows that nondimensionalization improves the efficiency of parameter recovery for physics-informed neural networks and for SINDy symbolic regression~\cite{brunton2016discovering}. There has been no work, to our knowledge, integrating universal differential equations~\cite{rackauckas2020universal} with the Buckingham $\Pi$ theorem.

\subsection{Buckingham $\Pi$ Theorem}\label{sec:buckingham}

The Buckingham $\Pi$ theorem is a fundamental tool in dimensional analysis, used to derive dimensionless quantities when the governing equation is not explicitly known~\cite{buckingham1914physically,barmeir2022fluidmechanics}. This theorem reduces the number of parameters in the original equation from \(n\) to \(n - k\), where \(n\) is the number of variables and \(k\) is the number of fundamental dimensions (such as length, mass, time, etc.). 

As an example, consider the equation for the potential energy difference due to gravity:
\[
U = G m_1 m_2 \left( \frac{1}{r_2} - \frac{1}{r_1} \right),
\]
where \(U\) is the potential energy difference between two points, with dimensions \([ML^2T^{-2}]\), \(G\) is the gravitational constant with dimensions \([M^{-1}L^3T^{-2}]\), \(m_1\) and \(m_2\) are the masses involved (dimensions \([M]\) each), and \(r_1\) and \(r_2\) are the distances from each mass to a point of interest (with dimensions \([L]\) each). Using the Buckingham \(\pi\) theorem, the number of variables in the equation can be reduced from six to three. The dimensional matrix \(\mathcal{D}\) representing the physical dimensions of each variable in the gravitational potential energy equation is structured as follows:
\[
\mathcal{D} = \bordermatrix{
& U & G & m_1 & m_2 & r_1 & r_2 \cr
\mathcal{M} & 1 & -1 & 1 & 1 & 0 & 0 \cr
\mathcal{L} & 2 & 3 & 0 & 0 & 1 & 1 \cr
\mathcal{T} & -2 & -2 & 0 & 0 & 0 & 0
}
\]
Three dimensionless variables (\(\pi\) groups) can be formed as follows:
\[
\pi_1 = \frac{U \, r_2}{G m_1^2} = \frac{[ML^{2}T^{-2}] \times [L]}{[M^{-1}L^{3}T^{-2}] \times [M]^2} = 1,
\]
\[
\pi_2 = \frac{m_2}{m_1} = \frac{[M]}{[M]} = 1,
\]
\[
\pi_3 = \frac{r_1}{r_2} = \frac{[L]}{[L]} = 1.
\]
This choice of \(\pi\) groups is just one possibility, and other combinations could be considered. One common method of deriving these dimensionless $\pi$ groups is the method of repeating variables~\cite{dumka2022implementation}.

To derive a functional relationship \( \pi_1 = f(\pi_2, \pi_3) \), we substitute for \( m_2 \) and \( r_1 \) in terms of \( \pi_2 \) and \( \pi_3 \) respectively, into the equation for \( U \):

\[
U = G m_1 m_2 \frac{r_1 - r_2}{r_1 r_2} = G m_1 m_2 \frac{(r_2 \pi_3) - r_2}{(r_2 \pi_3) r_2} = G m_1 m_2 \frac{r_2 (\pi_3 - 1)}{r_2^2 \pi_3}
\]

Simplifying further with \( m_2 = m_1 \pi_2 \):

\[
U = G m_1 (m_1 \pi_2) \frac{r_2 (\pi_3 - 1)}{r_2^2 \pi_3} = G m_1^2 \pi_2 \frac{\pi_3 - 1}{r_2 \pi_3}
\]

Substituting this expression for \( U \) into \( \pi_1 \):

\[
\pi_1 = \frac{U r_2}{G m_1^2} = \frac{G m_1^2 \pi_2 \frac{\pi_3 - 1}{r_2 \pi_3} r_2}{G m_1^2} = \frac{\pi_2 (\pi_3 - 1)}{\pi_3}
\]

Thus, the relationship \( \pi_1 = f(\pi_2, \pi_3) \) is established as:

\[
\pi_1 = \frac{\pi_2 (\pi_3 - 1)}{\pi_3}
\]

\subsection{Ipsen's Method of Dimensional Analysis}\label{sec:ipsen}

Ipsen's method for nondimensionalizing equations is an alternative method to Buckingham $\Pi$ which is used to nondimensionalize an equation~\cite{Ipsens_Method}. Ipsen's method also yields a set of $\pi$ terms, just like Buckingham $\Pi$, but this method provides flexibility to choose the dimensional terms precisely, which can be used to minimize the number of input variables for specific (unknown) expressions. The original form of the equation can subsequently be easily recovered from the dimensionless terms. Given the dependency of Ipsen's method on specific equations, we detail the application of Ipsen's method to two separate differential equations in the Methods section, and here we provide a general procedure. Another name for this method is the ``step-by-step'' method of nondimensionalization~\cite{clark2022alternate}.

Suppose that you have an equation
\[F(x_1,x_2\ldots,x_k)=0\]
where $x_i$ are variables present in the equation, with specific known dimensions. Ipsen's method of nondimensionalization is an iterative method where at each step, one can choose which dimension to eliminate. The output of the method is a set of dimensionless variables. Suppose that there are three variables, and $x_1 = [M], x_2 = [T], x_3 = [M^2T]$, and that
\[F(x_1,x_2,x_3)=0\]

One may let $x'_3 = x_3/x_1^2$ and $x'_1 = x_1/x_1$ which would now have dimensions of $[T]$ and 1 respectively. Then, one can substitute $x'_1,x'_3$ into the equation to yield

\[F(1,x_2,x_3/x^2_1)\]

Similarly, now dividing the second and third term by $x_2$ yields

\[F(1,1,x_3/x_2 x^2_1)\]

which shows us that the expression $F$ can be nondimensionalized with exactly one term $\pi = x_3/x_2 x^2_1$.

\subsection{Universal Physics-Informed Neural Networks}\label{sec:upinns}

Given data $\{x_i, t_i,u_i\}$ from a (partial or ordinary) differential equation, Universal Physics-Informed Neural Networks (UPINNs)~\cite{pmlr-v202-podina23a} can be used to find a representation of a hidden term using methods based on physics-informed neural networks~\cite{raissi2019physics}. Suppose that we have an ordinary differential equation of the following form with $m$ variables, where $u: t \rightarrow \mathds{R}^m$ and $P,Q: u,t \rightarrow \mathds{R}^m$:

\begin{equation}\label{eq:1}
    \frac{du}{dt} = P(u,t) + Q(u,t)
\end{equation}
where $P(u,t)$ is a known term and $Q(u,t)$ is an unknown term. $Q$, the unknown term, and $u$, the solution, are each represented by a fully connected neural network. Then, the following loss can be used to train the UPINN and find $Q(u,t)$:

\[\mathcal{L} = \mathcal{L}_{MSE} + \mathcal{L}_{ODE}\]

The MSE loss ensures that the output of the surrogate solution $U_{NN}$ (also represented by a neural network) adheres to the data $\{t_i,u_i\}$:

\[\mathcal{L}_{MSE} = \frac{1}{N}\sum_{i=1}^N (U_{NN}(t) - u_i)^2\]

As for the ODE loss, this loss minimizes the difference between the derivative of $U_{NN}(t)$ (autodifferentiated) and the right hand side of eq.~\ref{eq:1}.

\[\mathcal{L}_{ODE} = \frac{1}{K}\sum_{j=1}^K \Bigg(\left.\frac{dU_{NN} (t)}{dt}\right|_{t_j} - (P(U_{NN}(t_j),t_j) + Q_{NN}(U_{NN}(t_j),t_j))\Bigg)\]

The final output of the UPINN method is:
\begin{itemize}
    \item A list of points $(t_i,u_i)$ that represent the surrogate solution for the differential equation at time points $(t_i)$
    \item A list of points $(u_i,t_i,Q_i)$ that represent values of the hidden term for inputs $(u_i,t_i)$
\end{itemize}

This method has demonstrated strong performance in regimes characterized by low and noisy data~\cite{pmlr-v202-podina23a}, compared to Universal Differential Equations (UDEs)~\cite{rackauckas2020universal}, despite both approaches aiming to learn representations of hidden terms. Although here we demonstrate the application of the method to an ODE, this method can be adapted to a system of ODEs or a PDE as well. Finally, the relationship between the known term $P$ and the unknown term $Q$ may be multiplicative (or it may utilize a different binary operator).

\section{Methods}

We first illustrate how we integrated symbolic regression with the Buckingham $\Pi$ theorem to derive algebraic equations (Section~\ref{sec:methods_buckingham_pi}). We use this preliminary investigation to check if nondimensionalization in fact improves the accuracy of PySR~\cite{cranmerInterpretableMachineLearning2023}, a symbolic regression algorithm. Subsequently, to learn unknown components of differential equations, we employed a combination of PySR, dimensional analysis, and Universal Physics-Informed Neural Networks (UPINNs) (Section~\ref{sec:methods_ipsen}).

\subsection{Full Discovery of Algebraic Equations}\label{sec:methods_buckingham_pi}

We utilized the open-source model PySR~\cite{cranmerInterpretableMachineLearning2023} for symbolic regression to predict symbolic expressions of data from 7 different algebraic equations (Table~\ref{tab:equation_analysis}), each varying in complexity and variable count. We do this both for the traditional formulation of the algebraic equation, and after we nondimensionalize the data with the Buckingham $\Pi$ theorem. We include Coulomb's law, which is a special case because it reduces to exactly one $\pi$ term.  We do not need to run symbolic regression on an equation which reduces to exactly one $\pi$ term -- instead, the value of the $\pi$ term can be computed for each row of the data separately. In the case of noiseless data, the value of the $\pi$ term will be the same for all observations of the original variables. However, when there is noise, these values can be aggregated into a single value e.g. by taking the mean. Since the value of the $\pi$ term is a constant, the equation can be recovered exactly from this value by undoing the nondimensionalization.

PySR uses an evolutionary algorithm to construct and change expression trees that represent the full expression. Settings such as the functions used (e.g. $\sin, \cos, \log, \exp$), the maximum size of the expression, and the number of algorithm iterations are tunable. We set them to a maximum equation size of 20, number of iterations to 40, and the unary operators to: $\cos(x),\sin(x),\exp(x),\log(x),\sqrt{x},1/x$, and the binary operators of addition and multiplication. The settings were chosen based on a brief tutorial on PySR's GitHub page~\cite{cranmerInterpretableMachineLearning2023}, but we note that there could be significant sensitivity to these settings. 

First, we assessed the performance of PySR by evaluating its runtime and mean squared error on a train/test datasets of 10 datapoints each. Both training data and test data were generated for each equation (10 datapoints each), and only the training data was used to fit the PySR model. The test data was used for evaluation only. For each equation, data was generated with all independent variables being sampled uniformly between 0 and 1, except for the exponential decay example. In this case, the data ranges were set to uniform between $1\times 10^{-23}$ and $5\times 10^{-23}$ for the mass, and the Boltzmann constant $k_B=1.38\times 10^{-23}$. The reason for this is to keep the data within a realistic range. The gravitational constant $g$ was set to 9.81 m/s$^2$ in all cases. The gravitational constant $G = 6.67\times 10^{-11}$ m$^3$kg$^{-1}$s$^{-2}$. All other variables were sampled uniformly at random within the previously specified ranges.

Following this, we nondimensionalized the training data using the Buckingham $\Pi$ theorem and applied symbolic regression to the resulting dimensionless quantities, employing the same evaluation protocols. The objective of this comparative analysis was to determine the extent to which dimensional analysis enhances the performance of symbolic regression, specifically in terms of the accuracy of the equation fits. We note that while PySR was our chosen tool for symbolic regression, alternative algorithms could also be applied in this framework. Additionally, we note that while in our results, the nondimensionalized form of the equations is often not the most common form used by scientists (due to the process being handled automatically, and the specific set of dimensionless $\pi$ terms being chosen at random), the nondimensionalized equation can be easily converted to the original equation through direct substitution.

In terms of code contribution, we created an automated nondimensionalization framework based on Buckingham Py~\cite{dumka2022implementation}, which is a python-based library to implement automated derivation of dimensionless $\pi$ terms based on the Buckingham $\Pi$ theorem. The input is the original equation, and based on the dimensionless terms found by Buckingham Py, we are able to automatically nondimensionalize both an expression and its corresponding data. This was used to automatically process all of the algebraic equations.

\subsection{Discovery of Differential Equation Hidden Terms}
\label{sec:methods_ipsen}
% Ipsen's method with UPINNs and AI Feynman

For two distinct types of differential equations, first-order and second-order, we explored the effectiveness of Universal Physics-Informed Neural Networks (UPINNs)~\cite{pmlr-v202-podina23a} combined with dimensional analysis and PySR~\cite{cranmerInterpretableMachineLearning2023} in recovering the true equation under partial knowledge. First, we employed Ipsen's method to find the nondimensionalization that minimizes the number of input variables in the unknown term. Then, we trained a UPINN to recover the term under that partial nondimensionalization. Although Ipsen's method is not strictly necessary in this step, it offers some insight into the set of $\pi$ terms that minimizes the number of inputs to the unknown term. Finally, the form of the unknown term recovered by UPINNs can be passed to a symbolic regression algorithm such as AI Feynman~\cite{udrescu2020ai} or PySR~\cite{cranmerInterpretableMachineLearning2023}. Here, we use PySR as the symbolic regression algorithm. This nondimensionalization is designed to simplify the symbolic regression process, enabling the method to more effectively identify the true unknown term from its  UPINN approximation.

We now outline the differential equation setup and the steps that we perform, in detail, for three equations: the logistic growth DE, the rotating bead DE, and the extended Lotka-Volterra DE. Suppose we have a differential equation which has a known part $P$ and an unknown part $Q$, which may be vector-valued:

\[0 = P(a_1,a_2,\ldots,a_k) + Q(b_1,b_2,\ldots,b_l)\]

where $a_i$ and $b_j$ are variables appearing in the known and unknown terms respectively. We assume that either $P$ or $Q$ (or both) contain the spatial and/or temporal derivatives. These variables are assumed to have dimensions, and we assume that the dimensions are known, although some variables may be dimensionless. Additionally, we assume that we know which variables appear in $P$ and $Q$.

The steps that we perform for each differential equation are as follows:

\begin{enumerate}
    \item Note all the variables and their dimensions in the equation.
    \item Partially nondimensionalize the equation with guidance from Ipsen's method, and retrieve the set of $\pi$ terms. How to do this is shown in detail for each equation that we consider, but generally is a process dependent on the specific equation.
    \item Transform existing data to the nondimensional form (or, in the case of synthetic data, generate already dimensionless data), using the dimensionless $\pi$ terms from Step (2).
    \item Recover the unknown term $Q(\pi_1,\pi_2,\ldots,\pi_k)$ using UPINNs, as a function of only the $\pi$ terms that appear in it. This will be known in advance from the nondimensionalization.
    \item Apply symbolic regression to find a closed form for $Q(\pi_1,\pi_2,\ldots,\pi_k)$, or the unknown term as a function of the dimensionless variables.
    \item Optionally, the recovered equation in terms of dimensionless variables can be converted back to an equation with the original variables.
\end{enumerate}

Aside from following these steps, we also compare this method to not performing the nondimensionalization, and inferring the unknown term directly from the data. In this case, here are the steps: 

\begin{enumerate}
    \item Recover the unknown term $ Q(b_1,b_2,\ldots,b_l)$ using UPINNs as a function of the original variables that appear in $Q$.
    \item Apply symbolic regression to find a closed form for $ Q(b_1,b_2,\ldots,b_l)$, as a function of the input variables.
\end{enumerate}

The UPINN implementation follows \href{https://github.com/jayroxis/PINNs}{https://github.com/jayroxis/PINNs} and is in PyTorch~\cite{paszke2019pytorch}. The surrogate solution and unknown term was each represented by a multilayer perceptron with hidden layer sizes: $(20, 128, 128, 128, 128, 128, 128, 128)$, and tanh activation. We perform 1000 iterations of Adam and then several thousand of iterations of Limited Memory BFGS, until convergence. The PySR settings were the same as those used in the previous section.

We detail the application of our method to the extended logistic growth model (Section~\ref{sec:logistic_growth}), a first-order differential equation, and the rotating bead model (Section~\ref{sec:rotating_bead}), a second-order differential equation, and the Lotka-Volterra equation (Section~\ref{sec:lotka_volterra}) which is a system of two first-order ODEs.

\subsubsection{Extended Logistic Growth Model}\label{sec:logistic_growth}

To demonstrate the effectiveness of our method on a differential equation, we start with a variant of the logistic growth model, commonly used in computational biology~\cite{ali2021unraveling}:

\begin{equation}
\frac{dN}{dt} = rN \left(1 - \frac{N}{K}\right) + \frac{AN^2}{B^2 + N^2}
\end{equation}

where $N(t)$ are the number of cells over time, $t$ is time, $r,K,A,B$ being constant values. The dimensional matrix is:

\[
  D = \kbordermatrix{
    & N & r & K & A & B & t\\
    M & 1 & 0 & 1 & 1 & 1 & 0 \\
    T & 0 & -1 & 0 & -1 & 0 & 1 \\
  }
\]

where $M$ is the dimension mass, and $T$ is the dimension time. Here, $M$ represents an abstract population-count dimension rather than physical mass. This notation is introduced solely to facilitate dimensional analysis while preserving consistency among the quantities $N$, $K$, and $B$. For this problem, we are interested in recovering the true form of the unknown term $Q(A,N,B) = A N^2 / (B^2 + N^2)$. The rest of the differential equation is assumed to be known. The inputs to the function $Q$ are also assumed to be known.

Using the Buckingham $\Pi$ theorem to nondimensionalize this equation provides us with seven unique sets of $\pi$ values. However, it is initially unclear which set would allow the subsequent symbolic regression step to proceed with the highest accuracy in recovering the true unknown term. A $\pi$ set for this equation which minimizes the number of parameters and variables in the unknown term is the following:\\
\[\beta=\frac{Br}{A}\qquad \varepsilon=\frac{K}{B}\qquad  \tau=\frac{At}{B}\qquad \alpha=\frac{N}{B}\]
\vspace{5mm}
By algebraic manipulation, we can rewrite the original equation in terms of the non-dimensional parameters.

\begin{equation}
\frac{d \alpha}{d\tau} =\beta\alpha\left(1-\frac{\alpha}{\varepsilon}\right)+\frac{\alpha^2}{\alpha^2+1}
\end{equation}

When $Q$ is unknown, this equation takes the form of:

\begin{equation}
\frac{d\alpha}{d\tau} =\beta\alpha\left(1-\frac{\alpha}{\varepsilon}\right)+Q^*(\alpha)
\end{equation}

where $Q^*$ is the transformed unknown term.

By contrast, Ipsen's method offers a preliminary insight into the number of parameters each non-dimensional function will have, without the need to know what the unknown terms are. We express the ordinary differential equation as the sum of \(P\) and \(Q\), with the parameters shown in Equation~\ref{eq:logistic}:

\begin{equation}\label{eq:logistic}
0 = P(r, N, K, t) + Q(A, N, B)
\end{equation}

where $P = r N \left( 1 - N/K\right) - dN/dt$ and $Q$ is the unknown component $A N^2 / (B^2 + N^2)$. The goal is to minimize the number of parameters in the unknown function $Q$. The reason this is useful is because the fewer parameters $Q$ has, the simpler the function is to learn for UPINNs, and the simpler the subsequent equation discovery task is for symbolic regression. Ipsen’s method facilitates this by manipulating the number of parameters in each function without altering the overall behavior of the system, using dimensions as a scaling factor.

To illustrate, we show how to apply Ipsen's method in this example. 

First, we eliminate \(B\) (with dimension \([M]\)):

\begin{equation}
0 = P\left(r, \frac{N}{B}, \frac{K}{B}, t\right) + Q\left(\frac{A}{B}, \frac{N}{B},\frac{B}{B}\right)
\end{equation}

Next, we eliminate \(A/B\) (with dimension \([T]\)):

\begin{equation}
0 = P\left(\frac{rB}{A}, \frac{N}{B}, \frac{K}{B}, \frac{tA}{B}\right) + Q\left(\frac{AB}{AB}, \frac{N}{B},\frac{B}{B}\right)
\end{equation}

The method allows us to eliminate terms or variables ($B$ or $A/B$) from the unknown component by dividing all other variables by each term appearing in $Q$. This decreases the number of terms in $Q$ at each step. These steps are repeated until the remaining inputs to the functions $P$ and $Q$ are dimensionless. We note that here we can obtain the same set of dimensionless parameters as we obtained earlier with Buckingham $\Pi$.

Hence, the set of $\pi$ terms that minimizes the number of inputs to $Q$ is the above set,

\[\beta=\frac{Br}{A}\qquad \varepsilon=\frac{K}{B}\qquad\tau=\frac{At}{B}\qquad\alpha=\frac{N}{B}\]

We can nondimensionalize the partially known equation as follows:

\begin{enumerate}
    \item Nondimensionalize time and $N$: $\tau = At/B$, $\alpha=N/B$. So $dN/dt = d\alpha/d\tau \cdot A$. Substituting gives:
    \[0 = -A \frac{d\alpha}{d\tau} + rB\alpha \left( 1 - B\alpha/K 
    \right) + Q(A,N,B)\]
    \item Divide both sides by $A$, which appears in $Q$ and does not introduce new variables into it:
    \[0 = -\frac{d\alpha}{d\tau}-\frac{rB\alpha}{A}\left( 1 - B\alpha/K\right) + Q(A,N,B)/A\]
    \item 
    The only $\pi$ term that can appear in $Q^* = Q(A,N,B)/A$ must be $\alpha = N/B$. Hence, the other dimensionless terms in the equation are
    \[\gamma = rB/A, \varepsilon=K/B\]
    and they are in the known term only.
\end{enumerate}

We note that Ipsen's method is useful as a way to verify the process of the nondimensionalization, but it is not strictly necessary in order to perform the nondimensionalization.

Next, we need to create a dataset of tuples $\{A_i,N_i,B_i,\hat{Q}_i(A_i,N_i,B_i)\}$, where $\hat{Q}(A_i,N_i,B_i)$ is the unknown term inferred by the UPINN. In order to do this, we generate 10 solutions (from the original equation) with different parameters, and we set the parameters to be randomly sampled from a uniform distribution. Here are the ranges of each parameter: $r:(0.2,1.0)$; $K:(1,10)$; $A:(10,20)$; $B:(10,20)$. The initial condition is set to be $N_0=0.1$, with time ranging from 0 to 20. We generate 100 datapoints for each solution, and there is no added noise.

For each of these 10 solutions, we can construct a dataset $\{N_i,t_i\}$ with specific parameters $r,K,A,B$ which can be used to infer $\hat{Q}(N_i)$ using the UPINN. Concatenating each dataset and UPINN solution together, we can create the dataset $D_{dim} = \{A_i,N_i,B_i,\hat{Q}_i(A_i,N_i,B_i)\}$ on which PySR can be run, where $\hat{Q}$ is treated as a function of $A,N,B$. Without nondimensionalization, we attempt to recover $Q$ as a function of $A,N,B$. 

For the validation of our method, we nondimensionalize each of the ODE solutions with the above set of dimensionless terms to yield a dataset $\{\alpha_i,\tau_i\}$ with parameters $\gamma,\varepsilon$ as defined above, where $\{\alpha_i,\tau_i\}$ from each solution can be given to UPINNs to find $\hat{Q}^*(\alpha_i)$. We create a final dataset $D_{nondim} = \{\alpha_i,\hat{Q}^*(\alpha_i)\}$ by concatenating the UPINN unknown terms from each ODE solution, and recover $\hat{Q}^*(\alpha)$ as a function of $\alpha$ only. The results for this evaluation are detailed in Section~\ref{sec:results_nondim}.

\subsubsection{Rotating Bead Model}\label{sec:rotating_bead}

Now, we introduce a second-order differential equation which governs the displacement of a bead on a rotating hoop.

\begin{equation}
0 = -mr \frac{d^2\theta}{dt^2}  -b\frac{d\theta}{dt} - m g \sin \theta + mr \omega^2 \sin \theta \cos \theta
\end{equation}

We define the known term $P = - mr \frac{d^2\theta}{dt^2} -b\frac{d\theta}{dt}$ and the unknown term $Q = mr \omega^2 \sin \theta \cos \theta  - m g \sin \theta$. The dimensions of the variables here are:

\[
  D = \kbordermatrix{
    & m & r & t & \theta & \dot{\theta} & \ddot{\theta} & g & \omega & b\\
    M & 1 & 0 & 0 & 0 & 0 & 0 & 0 & 0 & 1\\
    L & 0 & 1 & 0 & 0 & 0 & 0 & 1 & 0 & 1\\
    T & 0 & 0 & 1 & 0 & -1 & -2 & -2 & -1 & -1\\
  }
\]

\begin{equation}
0 = P(m, r, b, t) + Q(m, g, r, \omega)
\end{equation} 

Eliminating \(m\) (dimension \([M]\)):

\begin{equation}
0 = P\left(\frac{m}{m}, r, \frac{b}{m}, t\right) + Q\left(\frac{m}{m},g, r, \omega\right)
\end{equation} 

Eliminating \(\omega\) (dimension \([T]\)):

\begin{equation}
0 = P\left(\frac{m}{m}, r, \frac{b}{m \omega}, t \omega\right) + Q\left(\frac{m}{m}, \frac{g}{\omega^2}, r, \frac{\omega}{\omega}\right)
\end{equation} 

Eliminating \(r\) (dimension \([L]\)):

\begin{equation}
0 = P\left(\frac{m}{m}, \frac{r}{r}, \frac{b}{m \omega r}, t \omega\right) + Q\left(\frac{m}{m}, \frac{g}{\omega^2 r}, \frac{r}{r}, \frac{\omega}{\omega}\right)
\end{equation} 

We have concluded that the set of dimensionless parameters that reduces the number of parameters in $Q$ is:

\[
\varepsilon = \frac{g}{\omega^2 r};\hspace{3mm}  
\tau = t\omega;\hspace{3mm}  
\gamma = \frac{b}{m \omega r}
\]

We can use this set of dimensionless parameters to nondimensionalize the equation. This set of parameters yields the following dimensionless form for the true underlying equation:

\begin{equation}\label{eq:nondim}
0 = -\frac{d^2\theta}{d\tau^2} -\gamma \frac{d\theta}{d\tau} - \varepsilon \sin \theta + \sin \theta \cos \theta
\end{equation}

In order to nondimensionalize the equation when it has an unknown term, we can follow the following procedure:

\begin{enumerate}
    \item Nondimensionalize time and/or the solution variable $\theta$. Here we use $\tau=t \omega$. Here $\theta$ is dimensionless so we don't need to nondimensionalize it. Substituting in dimensionless time yields:
    \[0 = -mr\omega^2\frac{d^2\theta}{d\tau^2} - b \omega \frac{d\theta}{d\tau} + Q(\theta, m, g, r, \omega)\]
    \item To reduce the number of variables in the known term and remove dimensions from the entire equation, divide through by a dimensional scaling factor that makes all terms in the equation dimensionless. Here, $m r \omega^2$ is a good choice because these are all variables that appear in $Q$ and hence we won't introduce any new variables into $Q$.
    \[0 = -\frac{d^2\theta}{d\tau^2} - \frac{b}{mr\omega}  \frac{d\theta}{d\tau} + Q(\theta, m, g, r, \omega)/(mr\omega^2)\]
    \item $Q$ can be nondimensionalized with $5-3=2$ terms:
    \[\theta, \varepsilon = \frac{g}{\omega^2 r}\]
    % In the logistic growth equation, you can nondimensionalize in a similar way.
    \item Read off (or calculate, if $\pi_2$ appears in the known term) the corresponding dimensionless quantities in the known term. Here, the remaining dimensionless quantity is
    \[\gamma = \frac{b}{mr\omega}\]
\end{enumerate}

Hence, Ipsen's method can be used to identify which exact sets of dimensionless variables will minimize the number of inputs to the unknown term, but the above procedure will allow us to actually perform the nondimensionalization.

For this differential equation, we proceed similarly to the logistic growth equation. To generate data, the original equation, with dimensions, is solved for different values of the parameters. We sample the parameters from a uniform distribution with the following ranges: $r:(1,100), m:(1,3),b:(1,3)$. We use the fixed parameters $g=9.8,\omega=1.0$ and fixed initial conditions $(\theta,\dot{\theta}=(3.14,-0.1))$, with time ranging from 0 to 20. Then, the differential equation is solved (100 evaluations per solution) for each of the 10 sets of parameters through time using \verb|scipy.odeint|~\cite{virtanen2020scipy}.

To construct $D_{dim}$, for each ODE solve, we employ UPINNs to recover $\hat{Q}(\theta)$ from observations $\{\theta_i,t\}$ with specific parameters $m,r,\omega,g,b$. Then, these estimates can be concatenated into a final dataset $D_{dim} = \{m_i,r_i,\theta_i,\hat{Q}_i(m_i,r_i,\theta_i)\}$. We do not need to include variables such as $b,\omega,g$ because they are either fixed or do not appear in $Q$. Then, we run PySR to recover a symbolic expression for $\hat{Q}(m,r,\theta)$.

In order to construct $D_{nondim}$, we first nondimensionalize each ODE solve to yield a dataset $\{\theta_i,\tau_i\}$ for each set of parameters $\gamma,\varepsilon$. Then, we use UPINNs to recover $\hat{Q}^*(\theta)$ for each ODE solve. Concatenating the results together, we create $D_{dim} = \{(\varepsilon_i,\theta_i,\hat{Q}^*(\theta_i,\varepsilon_i))\}$. Now, we can use PySR to recover $Q^*$ as a function of $\theta,\varepsilon$. We expect that symbolic regression would be able to recover the function $Q(\theta,\varepsilon) = -\varepsilon \sin \theta + \sin \theta \cos \theta$.

\subsection{Lotka--Volterra Model with a Hidden Saturation Correction}\label{sec:lotka_volterra}

To further demonstrate the scalability of our method to systems of ODEs, and to a case where the
optimal nondimensionalization renders the unknown term completely parameter-free, we consider
a predator-prey system of the form
\begin{align}
    \frac{du}{dt} &= \alpha u - \beta uv,\\
    \frac{dv}{dt} &= \delta uv - \gamma v + Q(u, v, \theta, h),
    \label{eq:lv_system}
\end{align}
where $u(t)$ and $v(t)$ denote the prey and predator populations, respectively, $\alpha$ is the
intrinsic prey growth rate, $\beta$ is the predation rate coefficient, $\delta$ is the predator
growth rate attributable to predation, and $\gamma$ is the predator death rate. The prey equation
is assumed to be fully known, whereas the predator equation contains an unknown nonlinear
correction term $Q$. The purpose of this example is to determine whether UPINNs can identify
a hidden biologically motivated mechanism from data when the remainder of the governing
equations is known.

The true hidden term is chosen to have a Holling type II functional form~\cite{holling1959some},
\begin{equation}
    Q(u, v, \theta, h) = -\frac{\theta\, uv}{h + u},
    \label{eq:holling}
\end{equation}
where $\theta$ controls the strength of the saturation effect and $h$ is the half-saturation
constant (the prey density at which the correction is half its maximum magnitude). Biologically,
this term represents a prey-dependent reduction in predator growth arising from saturation,
limited handling capacity, or imperfect conversion efficiency at high prey densities. Thus, $Q$
acts as a nonlinear correction to the classical Lotka--Volterra predator-growth term $\delta uv$
rather than replacing it.

We define the known and unknown parts as
\begin{align}
    0 &= P_1(u, v, \alpha, \beta, t),\\
    0 &= P_2(u, v, \delta, \gamma, t) + Q(u, v, \theta, h),
    \label{eq:lv_split}
\end{align}
where it is known that
\begin{align}
    P_1(u, v, \alpha, \beta, t) &= \alpha u - \beta uv - \frac{du}{dt}\\
    P_2(u, v, \delta, \gamma, t) &= \delta uv - \gamma v - \frac{dv}{dt}.
    \label{eq:lv_P}
\end{align}

Furthermore we define $\vec{P} = [P_1,P_2]$ and $\vec{Q} = [0,Q]$. In the following analysis we will use $Q$ and $\vec{Q}$ interchangeably. We can simplify the inputs and outputs of the known and unknown term as

\[\vec{0} = \vec{P}(u,v,\alpha,\beta,\delta,\gamma,t) + \vec{Q}(u,v,\theta,h)\]

The dimensions of all variables, using $[\mathrm{N}]$ for population count and $[\mathrm{T}]$
for time, are:

\[
  D = \kbordermatrix{
    & u & v & t & \alpha & \beta & \delta & \gamma & \theta & h \\
    N & 1 & 1 & 0 & 0 & -1 & -1 & 0 & 0 & 1 \\
    T & 0 & 0 & 1 & -1 & -1 & -1 & -1 & -1 & 0
  }
\]
Here $[N]$ denotes an abstract population-count dimension introduced to facilitate
dimensional analysis, in the same spirit as the logistic growth example. With $n = 9$ variables
and $p = 2$ independent dimensions, the Buckingham $\Pi$ theorem guarantees $9 - 2 = 7$
dimensionless groups.

\paragraph{Theoretical minimum number of inputs to the unknown term.}
Before applying Ipsen's method, we establish how far the unknown term can in principle be
reduced by examining $Q$'s own variables:
\[
  D_Q = \kbordermatrix{
    & u & v & \theta & h \\
    N & 1 & 1 & 0 & 1 \\
    T & 0 & 0 & -1 & 0
  }
\]
The rank of $D_Q$ is $2$ and $Q$ has $4$ variables, so the minimum number of dimensionless
groups for $Q$ alone is $4 - 2 = 2$. Crucially, $\theta$ is the only variable in $Q$ carrying
dimension $[\mathrm{T}^{-1}]$; it therefore cannot form a dimensionless ratio with any other
variable within $Q$. However, it can be fully absorbed into the definition of dimensionless time
at the level of the full equation, leaving $Q^{*}$ as a function of exactly the two groups
$x = u/h$ and $y = v/h$, with no free parameters.

\paragraph{Ipsen's method.}
We apply Ipsen's method to the full equation to identify the nondimensionalization that achieves
this minimum.

\medskip
\noindent\textit{Step 1 -- eliminate $[\mathrm{N}]$ using $h$:}
\begin{equation}
    0 = \vec{P}\!\left(\frac{u}{h},\, \frac{v}{h},\, \delta h,\, \gamma,\, \alpha, \beta h, t\right)
      + \vec{Q}\!\left(\frac{u}{h},\, \frac{v}{h},\, \theta,\, \frac{h}{h}\right).
    \label{eq:step1}
\end{equation}
After this step, $Q$ depends on the dimensionless variables $u/h$ and $v/h$, together with the
dimensional parameter $\theta\;[\mathrm{T}^{-1}]$.

\medskip
\noindent\textit{Step 2 -- eliminate $[\mathrm{T}]$ using $\theta$ (a parameter of $Q$ itself):}
\begin{equation}
    0 = \vec{P}\!\left(\frac{u}{h},\, \frac{v}{h},\, \frac{\delta h}{\theta},\,
        \frac{\gamma}{\theta},\, \frac{\alpha}{\theta}, \frac{\beta h}{\theta}, \theta t\right)
      + \vec{Q}\!\left(\frac{u}{h},\, \frac{v}{h},\, \frac{\theta}{\theta},\,
        \frac{h}{h}\right).
    \label{eq:step2}
\end{equation}
All arguments are now dimensionless and $Q$ depends only on $x = u/h$ and $y = v/h$,
achieving the theoretical minimum of two inputs. The key choice is using $\theta$---a parameter
belonging to $Q$---to define dimensionless time. Had we instead scaled time with $\alpha$ (the
prey growth rate, which does not appear in $Q$), a residual parameter ratio $\theta/\alpha$ would
remain inside the unknown term.

\paragraph{Dimensionless groups.}
The complete set of seven dimensionless $\pi$ terms consistent with the Buckingham $\Pi$
theorem is
\begin{equation}
    x = \frac{u}{h}, \qquad y = \frac{v}{h}, 
    \label{eq:pi_state}
\end{equation}
in the unknown term $Q$, together with five parameters that appear only in the known terms:
\begin{equation}
    \tau = \theta t,\qquad 
    a = \frac{\alpha}{\theta}, \qquad
    b    = \frac{\beta h}{\theta}, \qquad
    d    = \frac{\delta h}{\theta}, \qquad
    c    = \frac{\gamma}{\theta}.
    \label{eq:pi_params}
\end{equation}

\paragraph{Nondimensionalized system.}
Substituting $u = hx$, $v = hy$, $t = \tau/\theta$ into Eq.~\eqref{eq:lv_system} and using the
definitions in Eq.~\eqref{eq:pi_params} yields
\begin{align}
    h \theta \frac{dx}{d\tau} &= \alpha (hx)- \beta (h^2xy),\\
    h \theta \frac{dy}{d\tau} &= \delta (h^2xy) - \gamma (h y) + Q(x, y),
    \label{eq:lv_nondim1}
\end{align}

Dividing through by $\theta h$ gives:

\begin{align}
    \frac{dx}{d\tau} &= \frac{\alpha}{\theta} x- \frac{\beta h}{\theta} xy,\\
    \frac{dy}{d\tau} &= \frac{\delta h}{\theta} xy - \frac{\gamma}{\theta} y + Q^{*}(x, y),
    \label{eq:lv_nondim2}
\end{align}

where $Q^{*} = Q/(h\theta)$ is the rescaled unknown term. Substituting $a,b,c,d$ gives

\begin{align}
    \frac{dx}{d\tau} &= a x- b xy,\\
    \frac{dy}{d\tau} &= d xy - c y + Q^{*}(x, y),
    \label{eq:lv_nondim}
\end{align}
 Substituting the true hidden term
from Eq.~\eqref{eq:holling} gives
\begin{equation}
    Q^{*}(x, y)
        = \frac{Q(hx,\, hy,\, \theta,\, h)}{h\theta}
        = -\frac{\theta\,(hx)(hy)}{(h + hx)\,h\theta}
        = -\frac{xy}{1 + x}.
    \label{eq:Qstar}
\end{equation}
Thus $Q^{*}$ is completely parameter-free and depends only on the two dimensionless state
variables $x$ and $y$, confirming the theoretical minimum. The nondimensionalized predator
equation with the unknown hidden term is
\begin{equation}
    \frac{dy}{d\tau} = d\,xy - c\,y + Q^{*}(x, y).
    \label{eq:lv_pred_nondim}
\end{equation}
The goal is to recover $Q^{*}(x,y) = -xy/(1+x)$ from data. The original dimensional term is
obtained by undoing the rescaling:
\begin{equation}
    Q = h\theta\, Q^{*} = -\frac{\theta\, uv}{h + u}.
    \label{eq:Q_recover}
\end{equation}

% \paragraph{Nondimensionalized system.}
% Substituting $u = hx$, $v = hy$, $t = \tau/\theta$ into Eq.~\eqref{eq:lv_system} and using the
% definitions in Eq.~\eqref{eq:pi_params} yields
% %
% \begin{align}
%     \frac{dx}{d\tau} &= \frac{1}{\phi}\bigl(x - b\,xy\bigr),\\
%     \frac{dy}{d\tau} &= \frac{1}{\phi}\bigl(d\,xy - c\,y\bigr) + Q^{*}(x, y),
%     \label{eq:lv_nondim}
% \end{align}
% %
% where $Q^{*} = Q/(h\theta)$ is the rescaled unknown term. Substituting the true hidden term
% from Eq.~\eqref{eq:holling} gives
% %
% \begin{equation}
%     Q^{*}(x, y)
%         = \frac{Q(hx,\, hy,\, \theta,\, h)}{h\theta}
%         = -\frac{\theta\,(hx)(hy)}{(h + hx)\,h\theta}
%         = -\frac{xy}{1 + x}.
%     \label{eq:Qstar}
% \end{equation}
% %
% Thus $Q^{*}$ is completely parameter-free and depends only on the two dimensionless state
% variables $x$ and $y$, confirming the theoretical minimum. The nondimensionalized predator
% equation with the unknown hidden term is
% %
% \begin{equation}
%     \frac{dy}{d\tau} = \frac{1}{\phi}\bigl(d\,xy - c\,y\bigr) + Q^{*}(x, y).
%     \label{eq:lv_pred_nondim}
% \end{equation}
% %
% The goal is to recover $Q^{*}(x,y) = -xy/(1+x)$ from data. The original dimensional term is
% obtained by undoing the rescaling:
% %
% \begin{equation}
%     Q = h\theta\, Q^{*} = -\frac{\theta\, uv}{h + u}.
%     \label{eq:Q_recover}
% \end{equation}

\paragraph{Comparison with dimensional inputs.}
Without nondimensionalization we attempt to recover $Q$ as a function of $(u, v, \theta, h)$---four
dimensional inputs that vary across ODE solves. After nondimensionalization, the unknown term
is a function of $(x, y)$ only---two dimensionless inputs with no free parameters. This is the
most favourable reduction achieved across all examples in this paper: the logistic growth and
rotating bead examples both retain at least one free parameter in their nondimensionalized
unknown terms ($\alpha$ and $\varepsilon$, respectively), whereas here the unknown term is
entirely parameter-free. As a consequence, the dataset $\mathcal{D}_{\mathrm{nondim}}$ does not
need to carry any parameter columns alongside the UPINN estimates, which further simplifies
the symbolic regression input.

\paragraph{Data generation.}
We generate $10$ ODE solutions by sampling parameters uniformly at random from the following
ranges: $\alpha \in (0.5,\,1.5)$, $\beta \in (0.1,\,0.5)$, $\delta \in (0.1,\,0.3)$,
$\gamma \in (0.2,\,0.8)$, $\theta \in (0.5,\,2.0)$, $h \in (5,\,20)$. Initial conditions are
fixed at $u_0 = 10$, $v_0 = 5$, and each solution is integrated from $t = 0$ to $t = 30$ using
\texttt{scipy.odeint}~\cite{virtanen2020scipy}, yielding $100$ observations per solution. No noise is added.

\paragraph{Dataset construction.}
To construct $\mathcal{D}_{\mathrm{dim}}$, we apply UPINNs to each ODE solve with observations
$\{u_i, v_i, t_i\}$ and fixed known parameters $\alpha, \beta, \delta, \gamma$ to recover
$\hat{Q}(u_i, v_i)$ from the predator equation. Because $\theta$ and $h$ vary across solves,
they are included as additional columns so that PySR can search for expressions involving them.
Concatenating across all $10$ solves gives
\begin{equation*}
    \mathcal{D}_{\mathrm{dim}}
        = \bigl\{\,u_i,\; v_i,\; \theta_i,\; h_i,\;
          \hat{Q}_i(u_i, v_i, \theta_i, h_i)\bigr\}.
\end{equation*}
on which PySR is run to recover a symbolic expression for $Q(u, v, \theta, h)$.

For $\mathcal{D}_{\mathrm{nondim}}$, each ODE solve is first transformed via
$x = u/h$, $y = v/h$, $\tau = \theta t$ to yield a dimensionless dataset $\{x_i, y_i, \tau_i\}$.
Because the unknown term $Q^{*}$ depends on $x$ and $y$ only---with no residual parameters---no
additional columns are needed. UPINNs are used to recover $\hat{Q}^{*}(x_i, y_i)$ for each
solve, and the results are concatenated:
\begin{equation*}
    \mathcal{D}_{\mathrm{nondim}}
        = \bigl\{\,x_i,\; y_i,\;
          \hat{Q}^{*}_i(x_i, y_i)\bigr\}.
\end{equation*}
PySR is then run to recover a symbolic expression for $Q^{*}(x, y)$. We expect it to identify
$Q^{*}(x, y) = -xy/(1+x)$, a rational function in two variables with no free parameters, which
can be converted back to the original variables via
$Q = h\theta\, Q^{*} = -\theta uv/(h + u)$.

\section{Results}
\label{sec:results_nondim}

We detail the results of combining symbolic regression with the Buckingham $\Pi$ theorem to discover algebraic equations (Section~\ref{sec:buckingham_pi_ai_feynman}). We also show the results of learning unknown terms in three separate differential equations--the extended logistic growth model, the rotating bead model, and the extended Lotka-Volterra model--using UPINNs combined with partial nondimensionalization (Section~\ref{sec:ipsen_upinns}).

\subsection{PySR with Buckingham \texorpdfstring{$\pi$}{pi} theorem}
%\subsection{AI Feynman with Buckingham Pi}
\label{sec:buckingham_pi_ai_feynman}

\begin{figure}[h]
    \centering
     \subfigure[]{\includegraphics[width=0.85\textwidth]{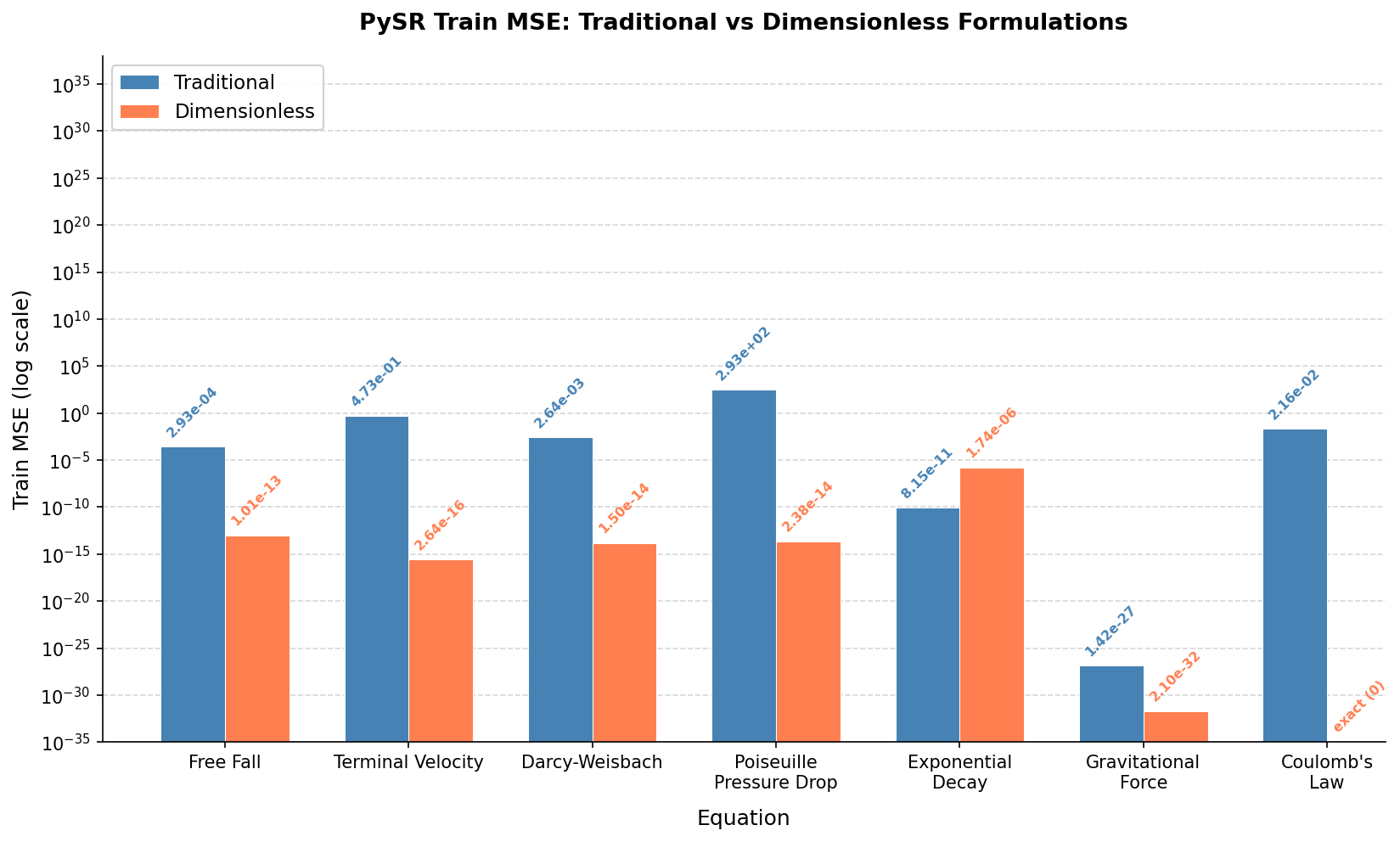}}
        % \caption{MSE: 0.002370}
        % \label{fig:linear}
    %--- Row 1 ---%
    \subfigure[]{\includegraphics[width=0.85\textwidth]{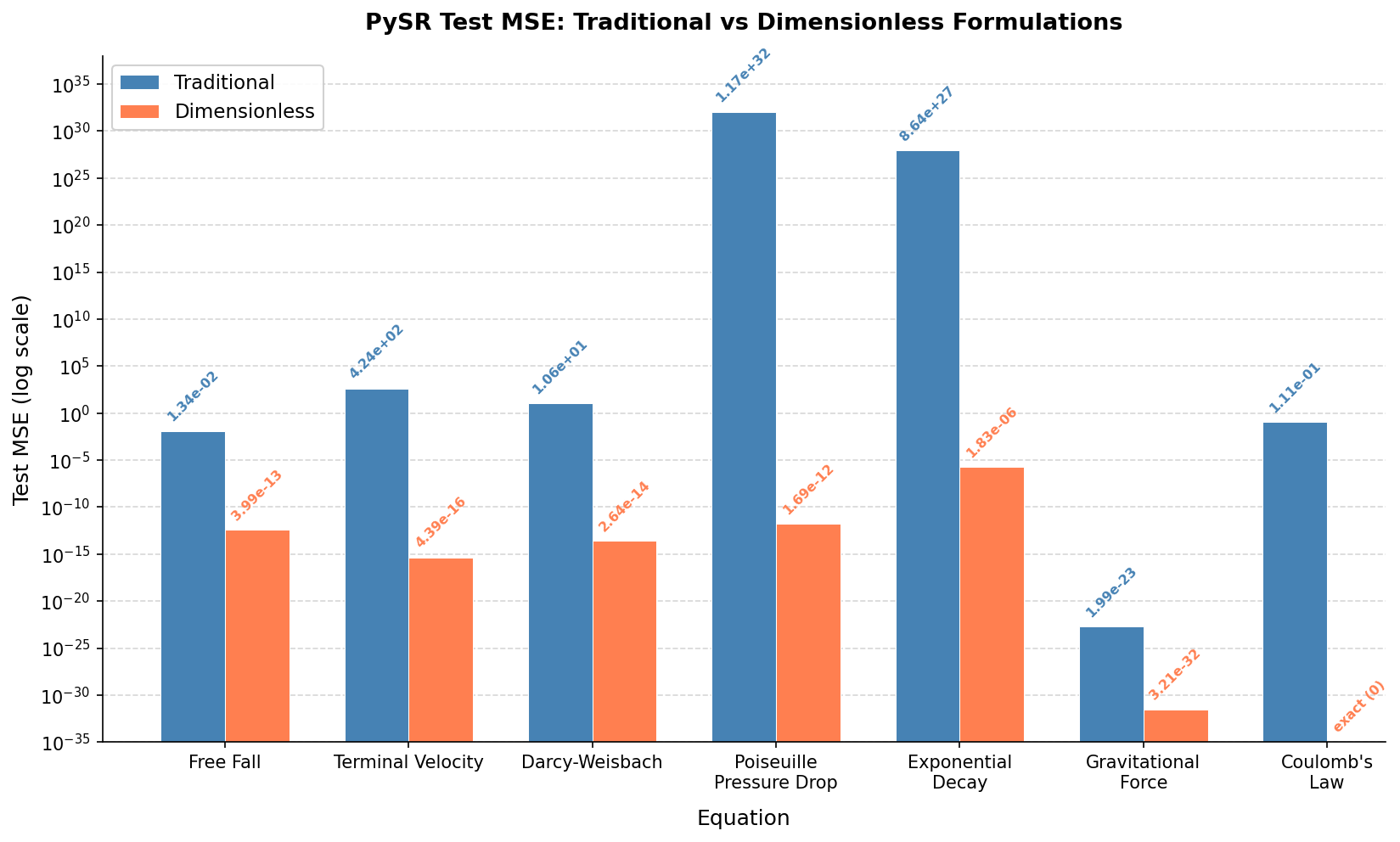}}

    \caption{Comparison of train and test mean squared error for each algebraic equation.}
    \label{fig:train_test_mse}
\end{figure}

In our experiment, we utilized 7 distinct algebraic equations to compare the solution time and error for finding both the original and dimensionally reduced equations. For each equation, we generate either 10 datapoints (Table~\ref{tab:equation_analysis}) or 100 datapoints (Table~\ref{tab:equation_analysis_100}), all noiseless. Each data entry in these tables contains the the target expressions for both the original and nondimensionalized formulation, the PySR-discovered expression, train and test MSE, and PySR equation discovery runtime. Figure~\ref{fig:train_test_mse} visualizes Table~\ref{tab:equation_analysis} and demonstrates the performance of the PySR model for each equation before and after dimensional analysis in terms of mean squared error on both the training set and the test set.

From Table~\ref{tab:equation_analysis} and Fig.~\ref{fig:train_test_mse} (10 datapoints), we can see that nondimensionalizing the equation yields better recovery of the true equation and better test MSE. For all equations, nondimensionalizing the data results in PySR being able to recover the true nondimensionalized equation exactly (disregarding small constants on the order of $10^{-7}$ and smaller) every time, except for exponential decay. For these equations, both training and test error is at most $10^{-10}$. In the case of exponential decay, the discovered equation still generalizes well to unseen test data, with a train and test MSE of $10^{-6}$. By contrast, the expressions found by PySR for the original equation forms are most of the time incorrect. The correct expression was recovered for only 1 of 7 equations, that of the gravitational force. The train and test MSE sufficiently small (less than $10^{-10}$) only for the gravitational force equation (original form). For the remainder of the equations in original form, the recovered expression does not generalize well to test data. Additionally, due to a consequence of the Buckingham $\Pi$ theorem, for Coulomb's law (last row), the nondimensionalization reduces the number of variables to a single $\pi$ term, which allows, in the case of noiseless data, exact recovery of the equation. The inference time does not significantly differ between the original and dimensionless versions of the same equation for all equations except the first two; it is possible that for larger datasets or for more complex equations, nondimensionalization would significantly decrease the time taken to converge. However, the decrease in computation time upon nondimensionalization for \textit{Free Fall} and \textit{Terminal Velocity} indicates that the computational load decreases.

We run the exact same experiments with 100 datapoints, and these results are shown in Table~\ref{tab:equation_analysis_100}. Compared to Table~\ref{tab:equation_analysis}, PySR always takes slightly longer to recover the original equation from the original data than the corresponding nondimensional equation from the transformed data. Additionally, in this regime, nondimensionalizing always decreases the runtime of PySR, showing promising time savings for higher numbers of datapoints. We see approximately the same train and test MSE results in both tables, confirming that the improvement in the ability of PySR to recover the true underlying equation is not dependent on the number of datapoints. As seen in Table~\ref{tab:equation_analysis_100}, nondimensionalizing enables PySR to recover the correct equation in all cases. For the original equations, adding more data essentially makes no difference to the accuracy of the discovered equations: \textit{Gravitational Force} and \textit{Free Fall} equations are identified correctly as before, with the addition of the symbolic expression identified for \textit{Exponential Decay} generalizing well, despite not matching the ground truth expression.

\begin{table}[h!]
\renewcommand{\arraystretch}{1.5}
\small
% \centering
\begin{tabular}{|p{3cm}|p{3.5cm}|p{3.5cm}|p{1.25cm}|p{1.25cm}|p{1.2cm}|}
\hline
\textbf{Equation Name} & \textbf{Original Equation} & \textbf{PySR-Deduced Equation} & \textbf{Train MSE} $(\downarrow)$ & \textbf{Test MSE} $(\downarrow)$ & \textbf{Runtime (s)} $(\downarrow)$ \\
\hline
\textbf{Free Fall} Traditional & $S = s_1 + vt + gt^2/2$ & $S = 0.83\cdot s_1\sqrt{\exp(v)} - t(4.32t - 0.20)$ & $2.925 \times 10^{-4}$ & $0.01336$ & 14.61\\
\hline
\textbf{Free Fall} Dimensionless & $\pi_1 = \pi_2 + \pi_3/2 + 1$ & $\pi_1 = \pi_2 + (1.65\times10^{-8})\pi_2 (\pi_2 - 0.612)\cos(\pi_3  + 0.762) + 0.50\pi_3 + 1.00$ & \boldmath$1.01\times 10^{-13}$ & \boldmath$3.99\times 10^{-13}$ & \textbf{5.36} \\
\hline
\hline
\textbf{Terminal Velocity} Traditional & $V = \sqrt{2mg/(rCA)}$ & $V = (-2A - 11.52m - 1.25)(0.309C + 0.309r\sin(A) - 1.0)/(C + r\sin(A))$ & $0.473$ & $424.5$ & 14.38 \\
\hline
\textbf{Terminal Velocity} Dimensionless & $\pi_1 = 2\pi_2/\pi_3$ & $\pi_1 = \pi_2(1.41\times 10^{-7}\pi_3 + 2.0)/\pi_3$ & \boldmath$2.64\times 10^{-16}$ & \boldmath$4.39\times 10^{-16}$ & \textbf{5.16} \\
\hline
\hline
\textbf{Darcy-Weisbach} Traditional& $P = f \cdot v^2/(2dg)$ & $P = (4.896f(1.477g - 1) + g(d - 3.123))(\log(\cos(v)) + 0.00808)/g$ & $2.64\times 10^{-3}$ & $10.56$ & \textbf{5.35} \\
\hline
\textbf{Darcy-Weisbach} Dimensionless& $\pi_1 = \pi_3/(2\pi_2)$ & $\pi_1 = \pi_3(1.06\times 10^{-7}\pi_3^2\cos(\pi_2) + 0.500)/\pi_2$ & \boldmath$1.50\times 10^{-14}$ & \boldmath$2.64\times 10^{-14}$ & 5.65 \\
\hline
\hline
\textbf{Poiseuille Pressure Drop} Traditional & $P = 128\mu LQ/(\pi D^4)$ & $P = (\mu\exp(L + 8.043\mu + 8.043\cos(D + 7.709)) + 7.391)/\mu$ & $292.7$ & $1.17\times 10^{32}$ & \textbf{5.20} \\
\hline
\textbf{Poiseuille Pressure Drop} Dimensionless & $\pi_1 = 0.0245\pi_2^4$ & $\pi_1 = 0.02454\pi_2(\pi_2 - 2.22\times 10^{-8})^2(\pi_2 + 1.66\times 10^{-7})$ & \boldmath$2.38\times 10^{-14}$ & \boldmath$1.69\times 10^{-12}$ & 5.81 \\
\hline
\hline
\textbf{Exponential Decay} Traditional & $n = n_0\exp(-mgx/(k_BT))$ & $n = 4.97\times 10^{-6}\exp(1/\sin(x)) + 9.42\times 10^{-6}\sin(0.0233/m) + 3.81\times 10^{-6}$ & \boldmath$8.15\times 10^{-11}$ & $8.64\times 10^{27}$ & \textbf{4.92} \\
\hline
\textbf{Exponential Decay} Dimensionless& $\pi_1 = 1/\log(\pi_2/\pi_3)$ & $\pi_1 = 0.205(\sqrt{\pi_3}(1.575\pi_2 + 1)/\pi_2)^{1/4} + 0.0262$ & $1.74\times 10^{-6}$ & \boldmath$1.83\times 10^{-6}$ & 5.45 \\
\hline
\hline
\textbf{Gravitational Force} Traditional & $F = -Gm_1m_2/r^2$ & $F = -6.669\times 10^{-11} \cdot m_1m_2/r^2$ & $1.42\times 10^{-27}$ & $1.99\times 10^{-23}$ & 5.07 \\
\hline
\textbf{Gravitational Force} Dimensionless & $\pi_1 = -\pi_2$ & $\pi_1 = -1.0\pi_2$ & \boldmath$2.10\times 10^{-32}$ & \boldmath$3.21\times 10^{-32}$ & \textbf{4.70} \\
\hline
\hline
\textbf{Coulomb's Law} Traditional & $F = q_1q_2/(4\pi \epsilon_0r^2)$ & $F = \log(\epsilon_0 + 0.209\exp(1/(r\cdot(5.81\cdot \epsilon_0 + 0.125))))$ & $0.0216$ & $0.1107$ & 14.17 \\
\hline
\textbf{Coulomb's Law} Dimensionless & $\pi_1 = 1/4\pi$ & $\pi_1 = 1/4\pi$ & \boldmath$0$ & \boldmath$0$ & -- \\
\hline
\end{tabular}
\vspace{1em}
\caption{Analysis of PySR-discovered algebraic equations with traditional and dimensionless forms, with 10 datapoints of noiseless data.}
\label{tab:equation_analysis}
\end{table}

\begin{table}[h!]
\renewcommand{\arraystretch}{1.5}
\small
% \centering
\begin{tabular}{|p{3cm}|p{3cm}|p{4.5cm}|p{1.25cm}|p{1.25cm}|p{1.2cm}|}
\hline
\textbf{Equation Name} & \textbf{Original Equation} & \textbf{PySR-Deduced Equation} & \textbf{Train MSE} $(\downarrow)$ & \textbf{Test MSE} $(\downarrow)$  & \textbf{Runtime (s)} $(\downarrow)$  \\
\hline
\textbf{Free Fall} Traditional & $S = s_1 + vt + gt^2/2$ & $S = s_1 - 4.9t^2 + t(v + 6.75\times10^{-8})$ & \boldmath$1.33\times10^{-15}$ & \boldmath$1.54\times10^{-15}$ & 15.32\\
\hline
\textbf{Free Fall} Dimensionless & $\pi_1 = \pi_2 + \pi_3/2 + 1$ & $\pi_1 = \pi_2 + 0.5\pi_3 + 1.000$ & $5.71\times10^{-14}$ & $5.71\times10^{-14}$ & \textbf{6.28} \\
\hline
\hline
\textbf{Terminal Velocity} Traditional & $V = \sqrt{2mg/(rCA)}$ & $V = \sqrt{m/(C\sin(r))}(4.220A + 1.034)/(A + 0.00817)$ & $2.15$ & $27.54$ & 6.19 \\
\hline
\textbf{Terminal Velocity} Dimensionless & $\pi_1 = 2\pi_2/\pi_3$ & $\pi_1 = 2.000\,\pi_2/\pi_3$ & \boldmath$1.74\times10^{-15}$ & \boldmath$2.72\times10^{-15}$ & \textbf{5.89} \\
\hline
\hline
\textbf{Darcy-Weisbach} Traditional& $P = f \cdot v^2/(2dg)$ & $P = -0.370fv(0.242d+1)(0.167g-1)/(dg)$ & $0.430$ & $212.8$ & 6.26 \\
\hline
\textbf{Darcy-Weisbach} Dimensionless& $\pi_1 = \pi_3/(2\pi_2)$ & $\pi_1 = 0.5\pi_3(\pi_2^{-2})^{1/2}$ & \boldmath$5.44\times10^{-28}$ & \boldmath$8.44\times10^{-27}$ & \textbf{6.01} \\
\hline
\hline
\textbf{Poiseuille Pressure Drop} Traditional & $P = 128\mu LQ/(\pi D^4)$ & $P = (108.32 - 1133.62D)\exp(L + 1.767\exp(\mu))/D^2$ & $5.98\times10^{10}$ & $3.20\times10^{22}$ & 7.48 \\
\hline
\textbf{Poiseuille Pressure Drop} Dimensionless & $\pi_1 = 0.0245\pi_2^4$ & $\pi_1 = \pi_2^3(0.02454\pi_2 - 4.48\times10^{-8}\sin(1.482\pi_2))$ & \boldmath$1.69\times10^{-5}$ & \boldmath$1.68\times10^{-8}$ & \textbf{6.45} \\
\hline
\hline
\textbf{Exponential Decay} Traditional & $n = n_0\exp(-mgx/(k_BT))$ & $n = n_0\sin(T)/(x^2\cos(2.168\exp(n_0)))$ & $4.39\times10^{-10}$ & $1.28\times10^{-8}$ & 6.43 \\
\hline
\textbf{Exponential Decay} Dimensionless& $\pi_1 = 1/\log(\pi_2/\pi_3)$ & $\pi_1 = 1/\log(\pi_2/\pi_3)$ & \boldmath$2.44\times10^{-35}$ & \boldmath$2.73\times10^{-35}$ &\textbf{ 6.17} \\
\hline
\hline
\textbf{Gravitational Force} Traditional & $F = -Gm_1m_2/r^2$ & $F = -6.035\times10^{-8}m_2^3\exp(\sqrt{m_1})\sin(\sqrt{m_1})/r$ & $9.41\times10^{-14}$ & $2.27\times10^{-14}$ & 6.04 \\
\hline
\textbf{Gravitational Force} Dimensionless & $\pi_1 = -\pi_2$ & $\pi_1 = -1.0\pi_2$ & \boldmath$1.12\times10^{-31}$ & \boldmath$8.43\times10^{-31}$ & \textbf{5.56} \\
\hline
\end{tabular}
\vspace{1em}
\caption{Analysis of PySR-discovered algebraic equations with traditional and dimensionless forms, 100 noiseless datapoints per equation.}
\label{tab:equation_analysis_100}
\end{table}

\subsection{Symbolic Regression with Ipsen's method and UPINNs}
\label{sec:ipsen_upinns}
We tested our method on three differential equations, i.e. the extended logistic growth equation (both noisy and noiseless data), the rotating bead equation of motion, and the extended Lotka-Volterra system of ODEs, comparing the results of discovered expression through UPINNs and PySR to the true equation.

\subsubsection{Logistic growth ODE: noisy and noiseless data}

\begin{figure}[h]
    \centering
     \subfigure[]{\includegraphics[width=0.45\textwidth]{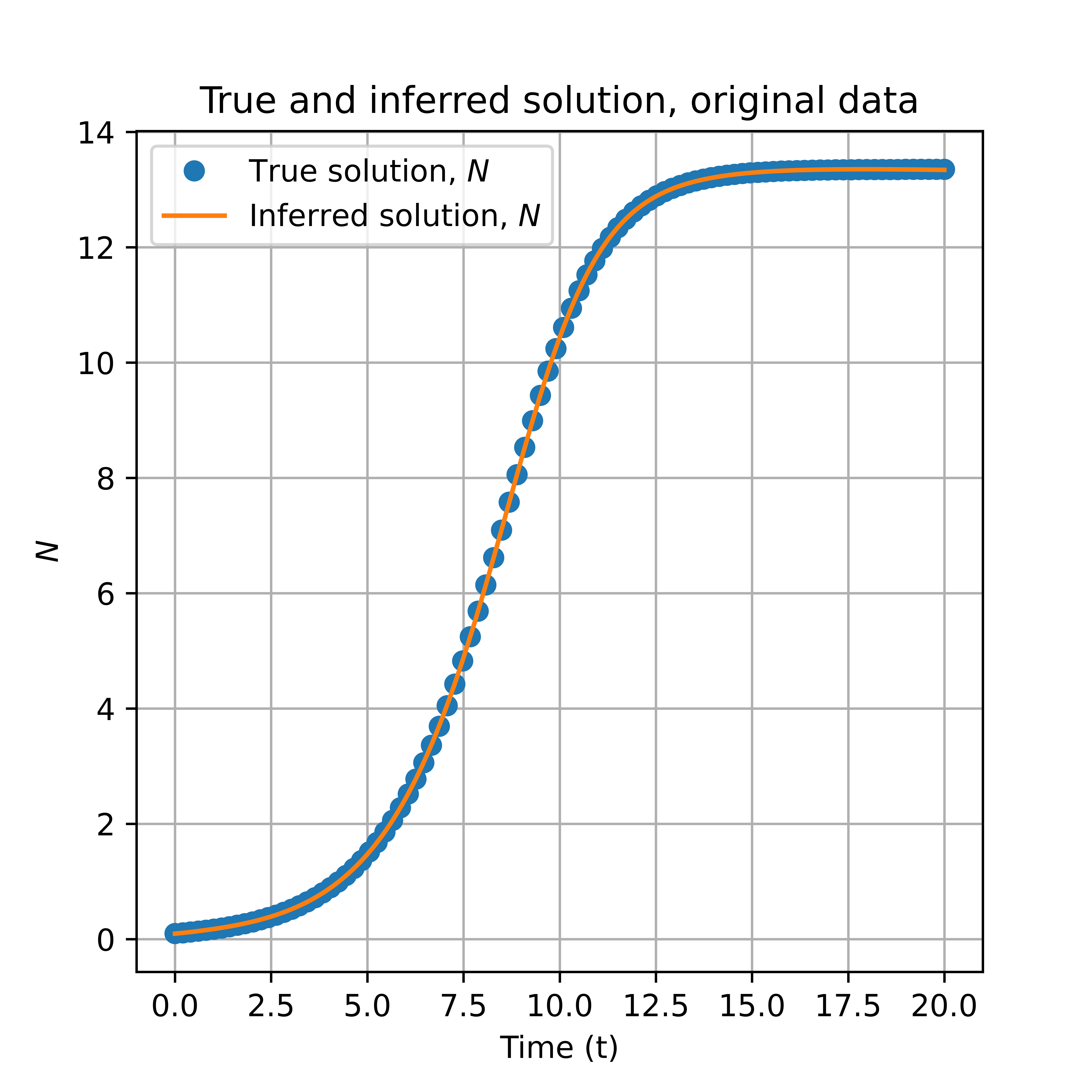}}
        % \caption{MSE: 0.002370}
        % \label{fig:linear}
    %--- Row 1 ---%
    \subfigure[]{\includegraphics[width=0.45\textwidth]{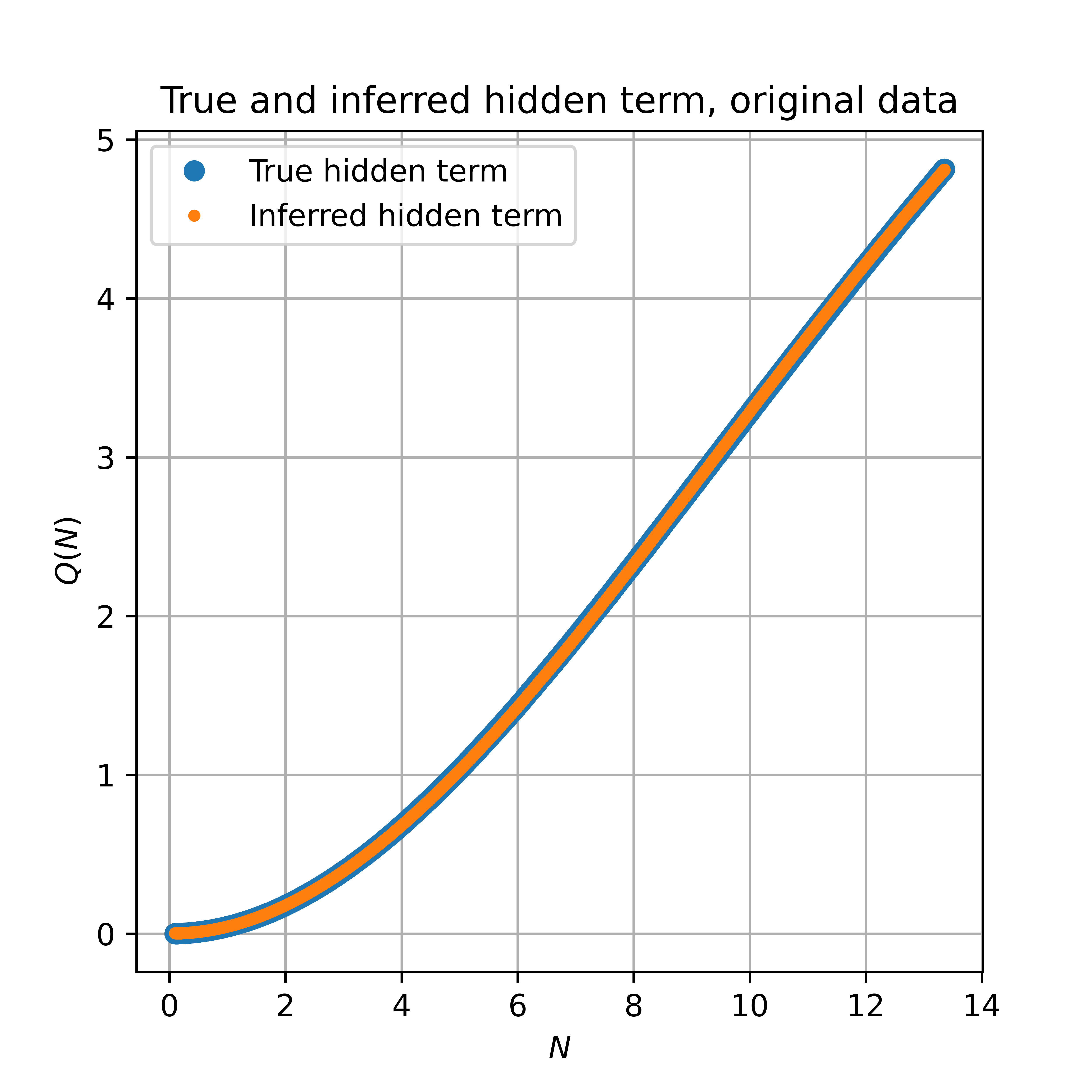}}
    \subfigure[]{\includegraphics[width=0.45\textwidth]{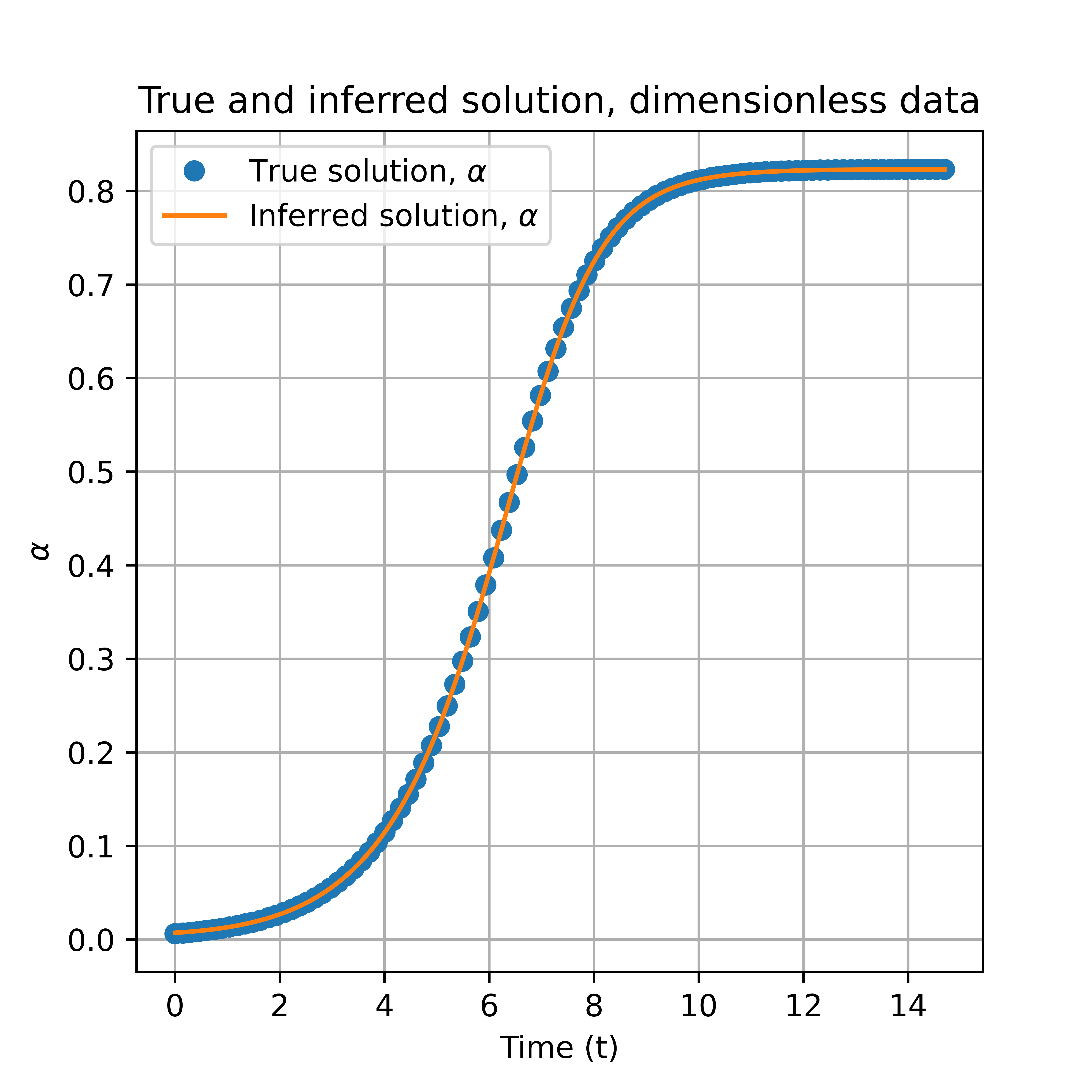}}
        % \caption{MSE: 0.002370}
        % \label{fig:linear}
    %--- Row 1 ---%
    \subfigure[]{\includegraphics[width=0.45\textwidth]{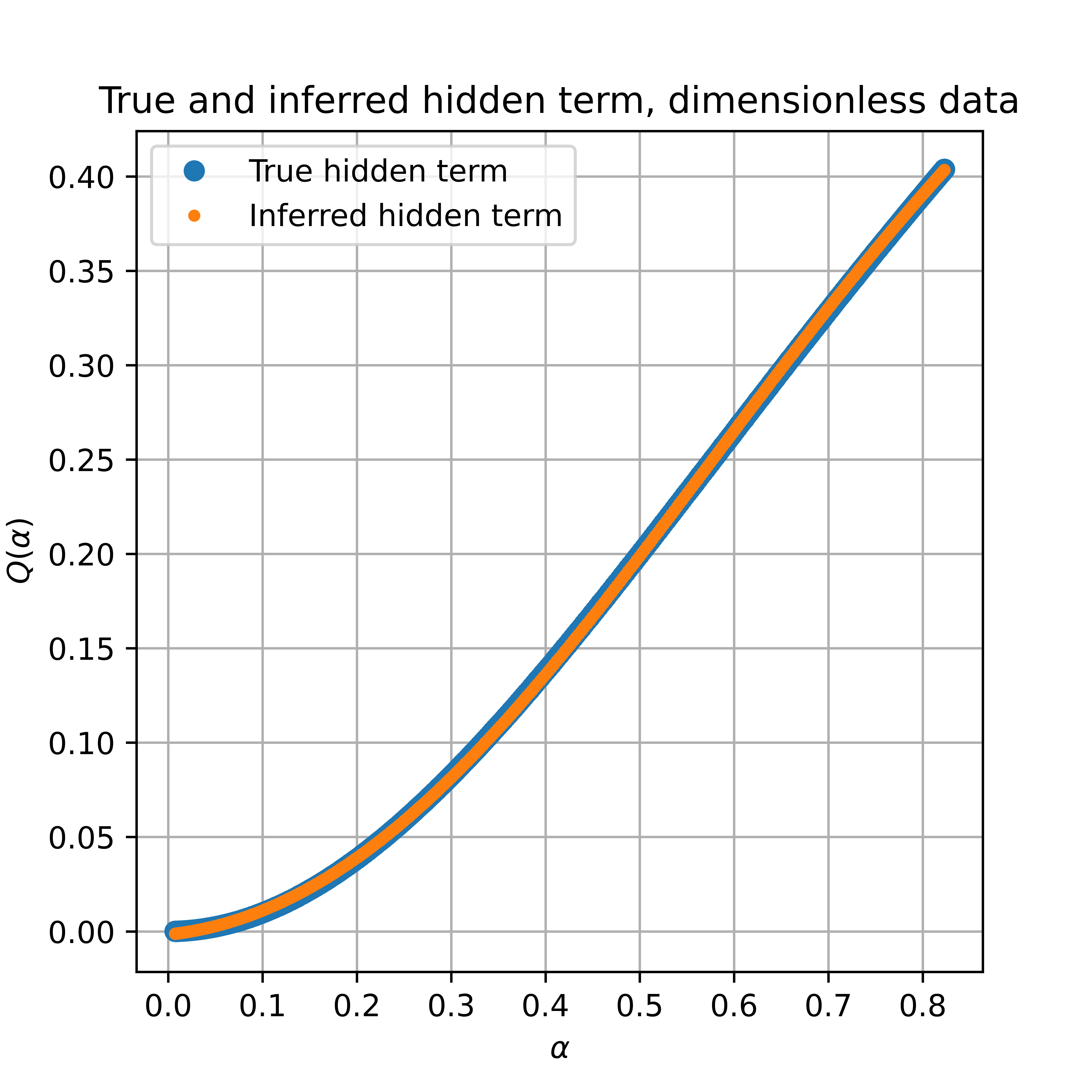}}

    \caption{Solution and unknown term fit for the logistic equation, for one set of parameters. UPINNs do very well in recovering the unknown term in both cases of the original and the dimensionless equation. (a) and (b) are the original equation, whereas (c) and (d) are the fits for the dimensionless equation.}
    \label{fig:logistic_growth}
\end{figure}

For the extended logistic growth equation, we test our workflow on nondimensionalized and original data as per the methods. We plot the surrogate solution and the unknown term as recovered by the UPINN in Figure~\ref{fig:logistic_growth}. We can see that for the dimensionless and original data, the UPINN is very good at recovering the unknown term in both scenarios. Table~\ref{tab:logg_mse} shows the hidden term MSE for all the different sampled parameters, which is generally below $10^{-4}$ for both dimensionless and original equations. This is computed by comparing the recovered hidden term with the true form of the term as computed from the variables in the surrogate solution. Table~\ref{tab:logg_symreg} shows the recovered equations for the original data and the nondimensionalized data. We note that for the original data, none of the proposed candidate expressions are correct, or have a loss less than 0.82. For the nondimensionalized data, PySR is able to recover the true expression exactly (row 6). To investigate whether a larger amount of data could help the unknown term be recovered more precisely for the original data, we show the results of repeating the experiment with 50 ODE solves (50 random sets of parameters) in Table~\ref{tab:logistic_growth_pysr_50}. Although the lowest loss is now 0.71, the true equation is still not recovered.

\begin{table}[htbp]
\centering
\begin{tabular}{cccccccc}
\toprule
$r$ & $K$ & $A$ & $B$ & $\gamma$ & $\varepsilon$ & MSE (orig.) & MSE (nondim.) \\
\midrule
0.5502 & 8.0682 & 11.9152 & 16.2211 & 0.7490 & 2.0105 & $8.38 \times 10^{-6}$ & \boldmath$6.54 \times 10^{-7}$ \\
0.4212 & 8.2168 & 17.7998 & 12.7259 & 0.3011 & 1.5488 & $1.51 \times 10^{-5}$ & \boldmath$7.51 \times 10^{-8}$ \\
0.4863 & 5.5090 & 19.5814 & 18.7593 & 0.4658 & 3.4052 & $1.99 \times 10^{-5}$ & \boldmath$1.16 \times 10^{-7}$ \\
0.4962 & 6.0508 & 16.8346 & 17.1270 & 0.5048 & 2.8306 & $1.86 \times 10^{-6}$ & \boldmath$2.92 \times 10^{-7}$ \\
0.8183 & 8.9438 & 15.0308 & 10.1377 & 0.5519 & 1.1335 & $2.71 \times 10^{-2}$ & \boldmath$2.19 \times 10^{-5}$ \\
0.2603 & 4.3194 & 13.6489 & 16.1540 & 0.3081 & 3.7398 & $2.52 \times 10^{-8}$ & \boldmath$1.00 \times 10^{-7}$ \\
0.5178 & 8.0986 & 19.3314 & 16.5138 & 0.4423 & 2.0391 & $4.11 \times 10^{-6}$ & \boldmath$1.64 \times 10^{-7}$ \\
0.8953 & 4.9256 & 13.1684 & 15.6810 & 1.0661 & 3.1836 & $6.19 \times 10^{-7}$ & \boldmath$6.69 \times 10^{-8}$ \\
0.7634 & 7.3412 & 18.0215 & 11.4377 & 0.4845 & 1.5580 & $2.71 \times 10^{-6}$ & \boldmath$6.99 \times 10^{-7}$ \\
0.5537 & 9.1838 & 12.1879 & 19.2487 & 0.8745 & 2.0959 & \boldmath$7.73 \times 10^{-7}$ & $6.75 \times 10^{-6}$ \\
\bottomrule
\end{tabular}
\vspace{1em}
\caption{Parameter values and MSE for original and nondimensionalized fits of the unknown term for the logistic growth equation.}
\label{tab:logg_mse}
\end{table}

\begin{table}[htbp]
\centering
\begin{tabular}{rcl}
\toprule
Complexity & Loss & Discovered equation (ground truth: $AN^2/(B^2+N^2)$)\\
\midrule
1  & 18.84   & $4.08144$ \\
2  & 12.89   & $\sqrt{N}$ \\
3  & 2.75   & $N \cdot 0.554923$ \\
4  & 2.75    & $\exp(N \cdot 0.1348798)$ \\
5  & 1.79   & $(A \cdot 0.03401371) \cdot N$ \\
6  & 1.04   & $N \cdot \text{inv}(B \cdot 0.13509956)$ \\
7  & 0.99  & $N \cdot (\cos(\sqrt{B}) \cdot -0.7250628)$ \\
8  & 0.87  & $\sin(\sin(\log(B) \cdot 0.9430373)) \cdot N$ \\
10 & 0.81  & $(\sin(\sin(0.9430345 \cdot \log(B))) \cdot N) + (-0.23878135)$ \\
\midrule
Complexity & Loss & Discovered equation (ground truth: $\alpha^2/(\alpha^2+1)$)\\
\midrule
1 & $6.49\times 10^{-2}$ & $0.25141922$ \\
2 & $5.78\times 10^{-2}$ & $\sin(\alpha)$ \\
3 & $9.32\times 10^{-4}$ & $\alpha \cdot 0.45889705$ \\
5 & $6.02\times 10^{-4}$ & $(\alpha -0.05566975) \cdot 0.48454857$ \\
6 & $5.33\times 10^{-4}$ & $\sin((\alpha \cdot 0.53107023) -0.03838172)$ \\
\textbf{7} & \boldmath$3.08\times 10^{-6}$ & $\boldsymbol{\alpha \cdot \text{inv}(\text{inv}(\alpha) + \alpha)}$ \\
9 & $3.08 \times 10^{-6}$ & $\alpha \cdot \text{inv}(\alpha + (\text{inv}(\alpha) \cdot 1.0001483))$ \\
\bottomrule
\end{tabular}
\vspace{1em}
\caption{Symbolic expressions returned by PySR for the extended logistic growth model (top: original data, and bottom: nondimensionalized data, 10 ODE solves for each). \textit{inv} represents the function $1/x$. The true discovered equation is bolded, and is algebraically equivalent to the true unknown term, $\alpha^2/(1+\alpha^2)$.}
\label{tab:logg_symreg}
\end{table}

\begin{table}[htbp]
\centering
\begin{tabular}{rcl}
\toprule
Complexity & Loss & Discovered equation (ground truth: $AN^2/(B^2+N^2)$) \\
\midrule
1 & $15.01$  & $3.293288$ \\
2 & $14.87$   & $\sqrt{A}$ \\
3 & $2.14$   & $N \cdot 0.52549046$ \\
5 & $1.69$   & $N \cdot (N \cdot 0.03511109)$ \\
6 & $1.36$   & $N \cdot (\text{inv}(B) \cdot 6.700879)$ \\
7 & $1.12$   & $(N \cdot 0.017628053) \cdot (A + N)$ \\
8 & $0.79$  & $N \cdot (\text{inv}(B) \cdot (A \cdot 0.4347375))$ \\
9 & $0.71$   & $N \cdot (A \cdot ((B \cdot -0.0022081388) + 0.06363801))$ \\
\bottomrule
\end{tabular}
\vspace{1em}
\caption{Symbolic expressions returned by PySR for the logistic growth DE (original equation, data from 50 ODE solves).}
\label{tab:logistic_growth_pysr_50}
\end{table}

We also test our method on data from the logistic equation with noise level $\epsilon=0.05$ and $\epsilon=0.1$. We add noise proportionally to the mean as follows:

\[x_{noisy} = x + \bar{x}\cdot \mathcal{N}(0,1)\cdot \epsilon\]

\begin{figure}[h]
    \centering
     \subfigure[]{\includegraphics[width=0.45\textwidth]{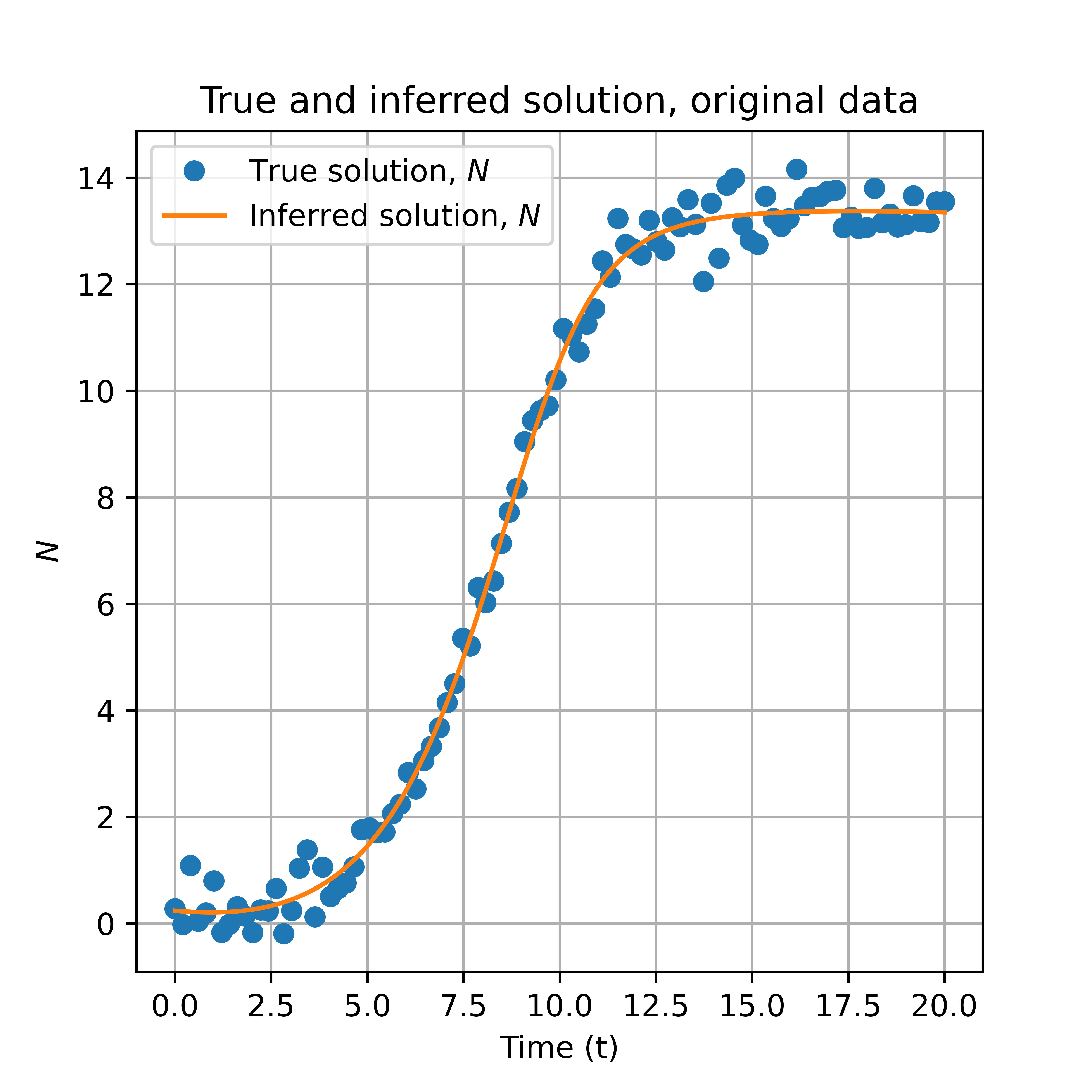}}
        % \caption{MSE: 0.002370}
        % \label{fig:linear}
    %--- Row 1 ---%
    \subfigure[]{\includegraphics[width=0.45\textwidth]{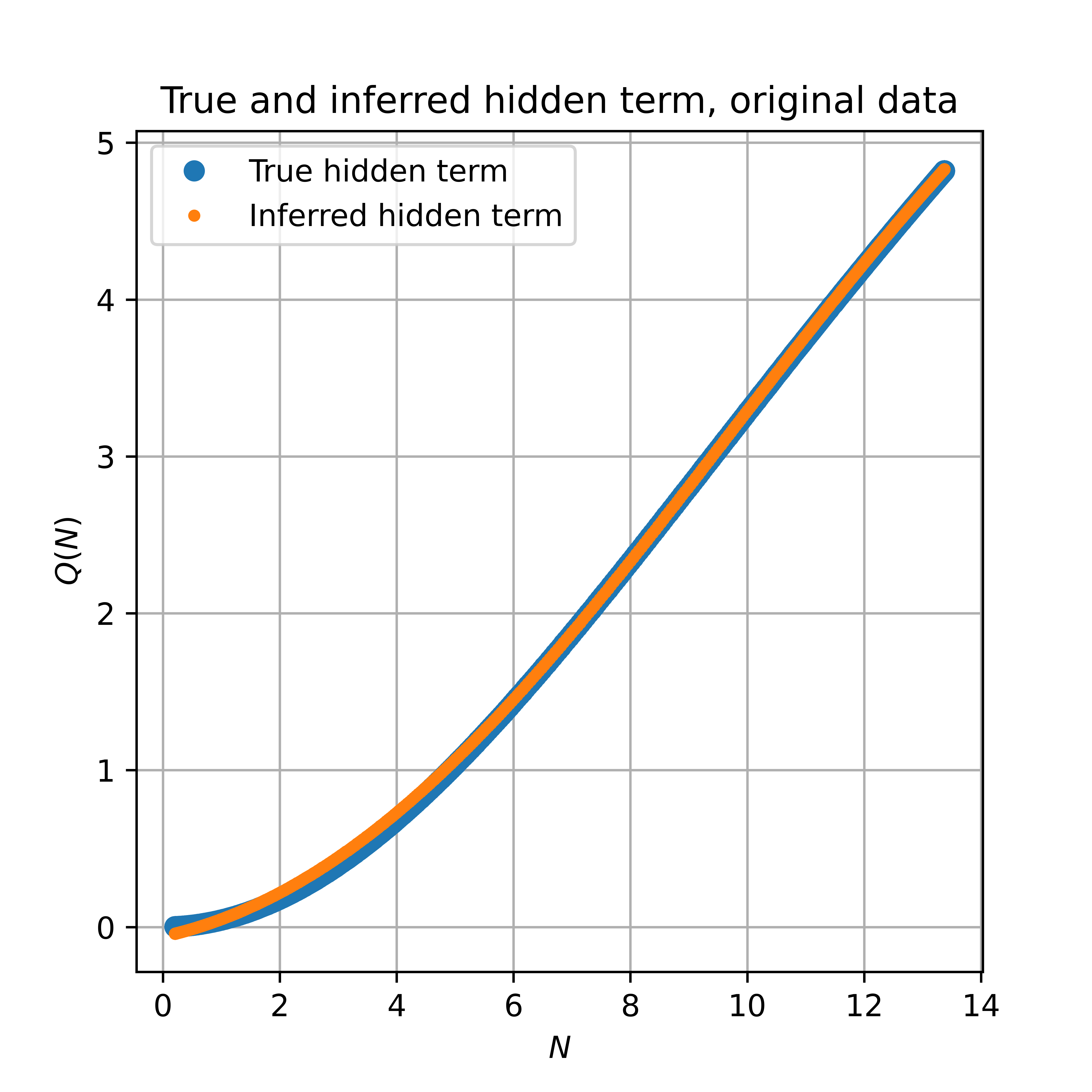}}
    \subfigure[]{\includegraphics[width=0.45\textwidth]{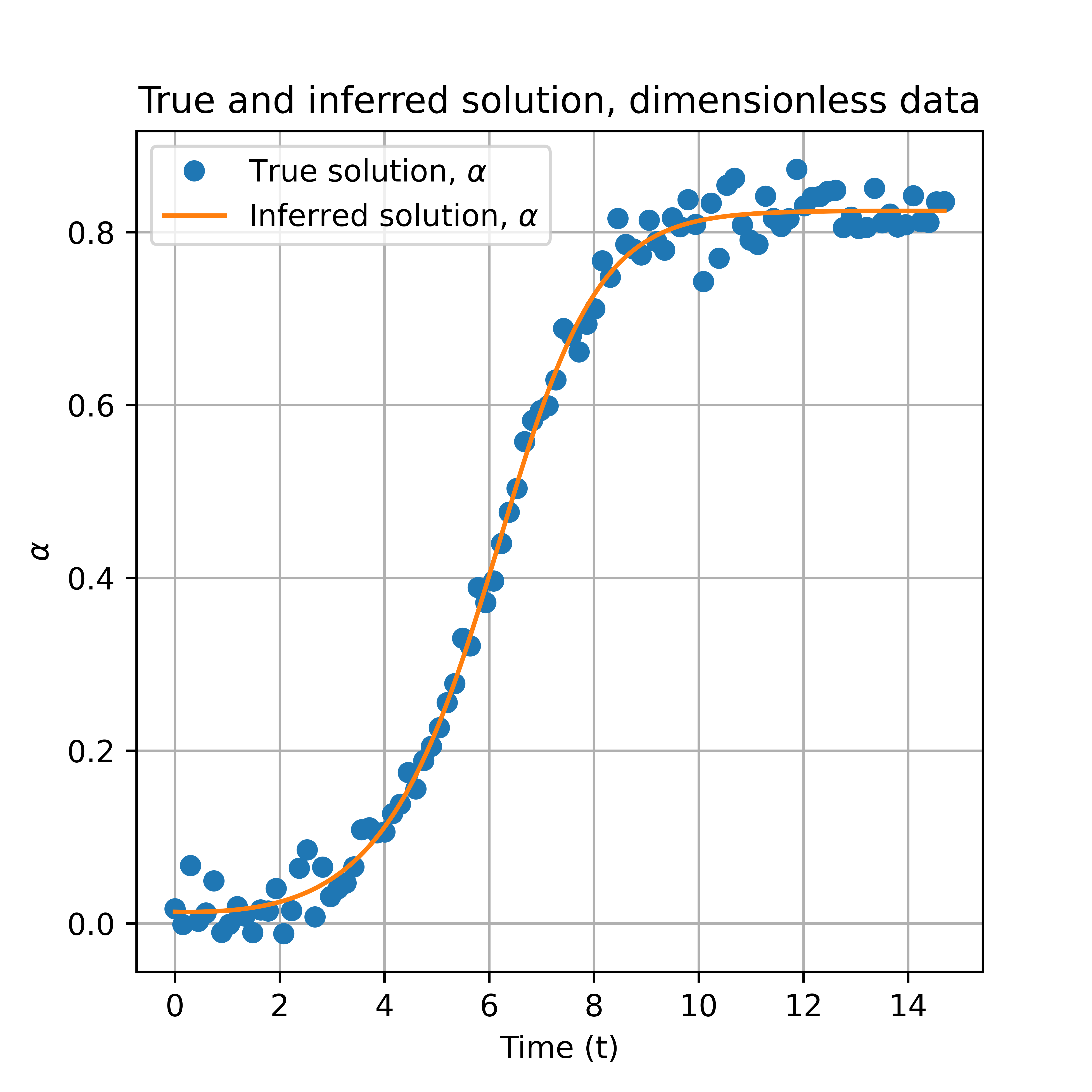}}
        % \caption{MSE: 0.002370}
        % \label{fig:linear}
    %--- Row 1 ---%
    \subfigure[]{\includegraphics[width=0.45\textwidth]{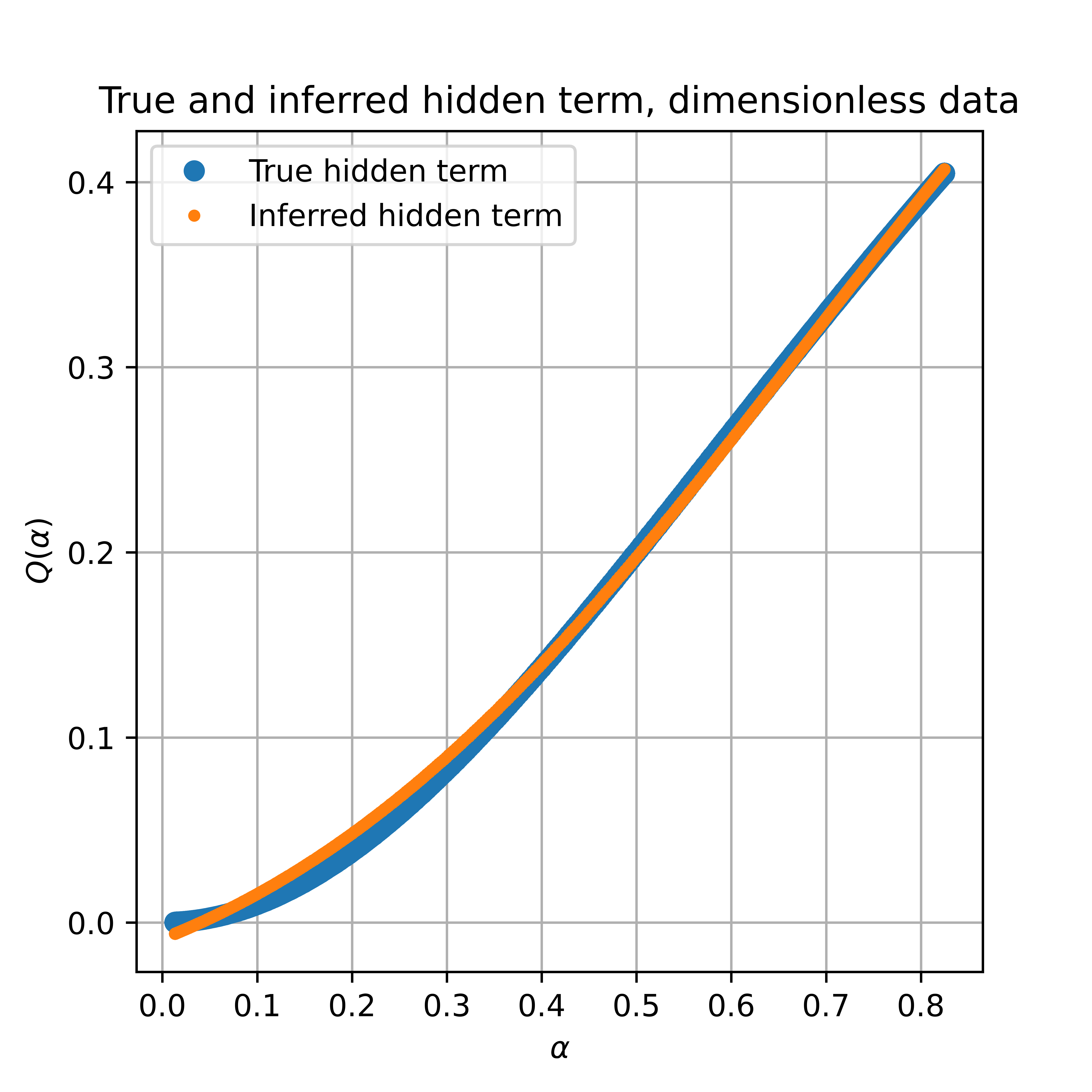}}

    \caption{Solution and hidden term fit for the logistic equation (noise level 0.05), for one set of parameters. (a) and (b) are the original equation, whereas (c) and (d) are the fits for the dimensionless equation.}
    \label{fig:logistic_growth_noise5}
\end{figure}

\begin{figure}[h]
    \centering
     \subfigure[]{\includegraphics[width=0.45\textwidth]{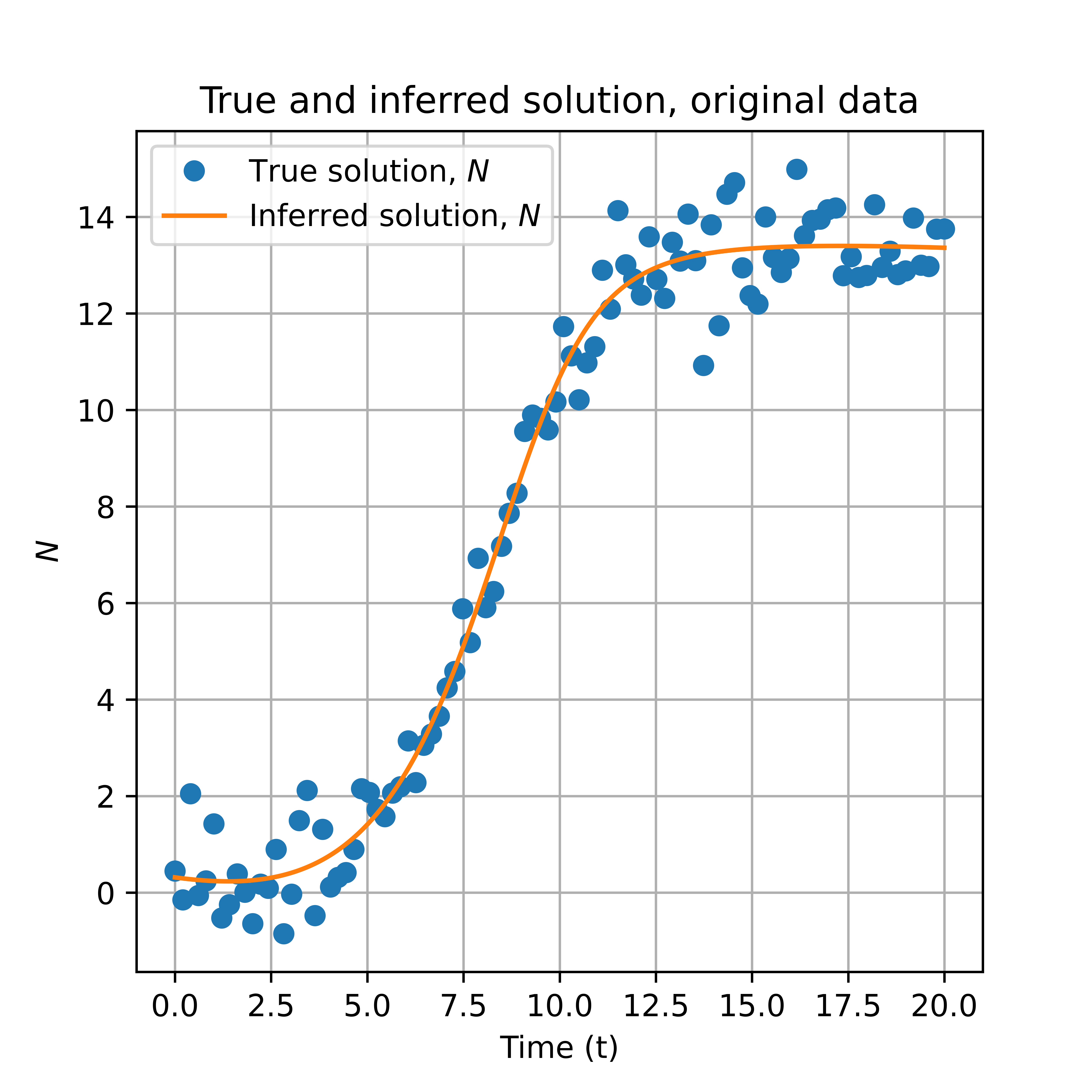}}
        % \caption{MSE: 0.002370}
        % \label{fig:linear}
    %--- Row 1 ---%
    \subfigure[]{\includegraphics[width=0.45\textwidth]{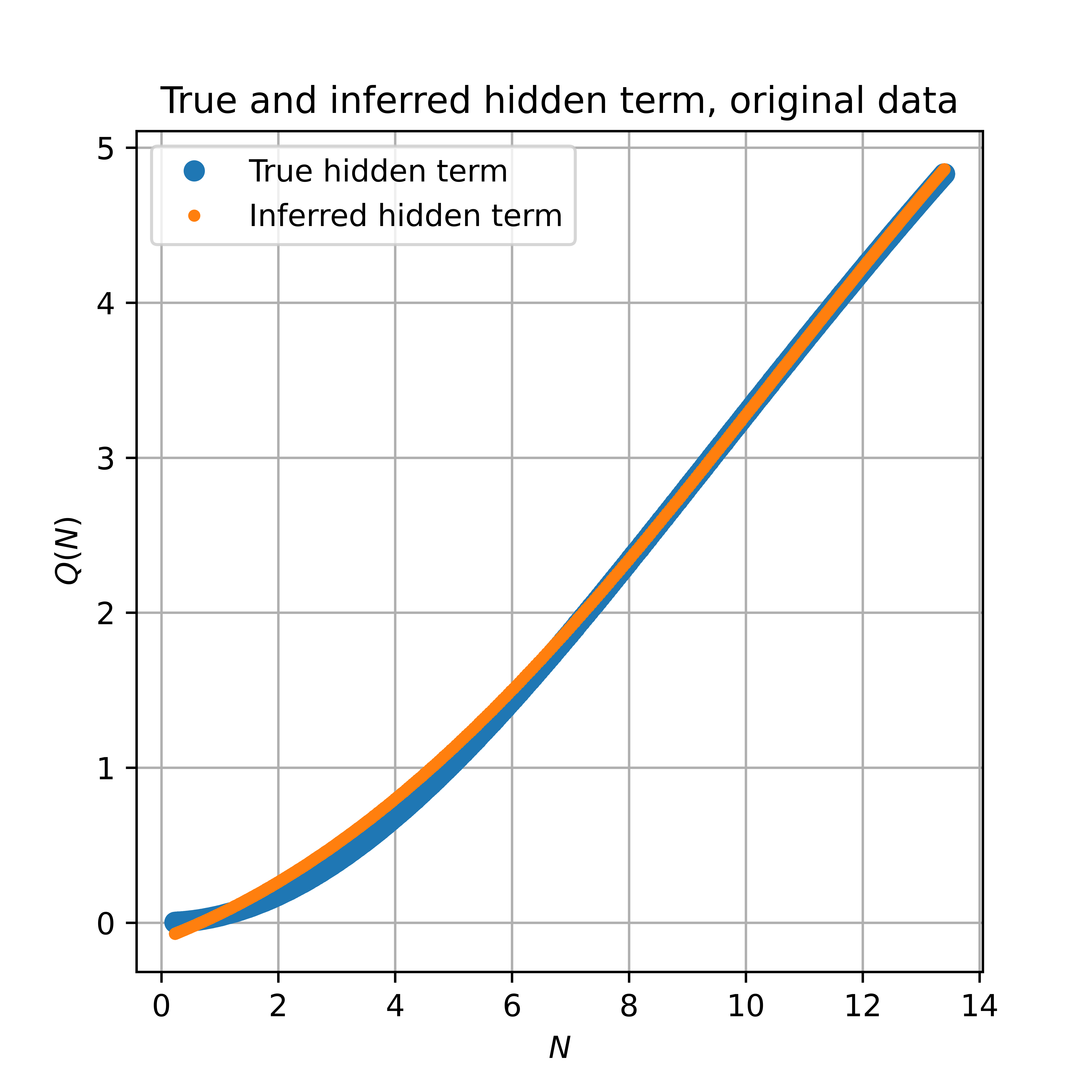}}
    \subfigure[]{\includegraphics[width=0.45\textwidth]{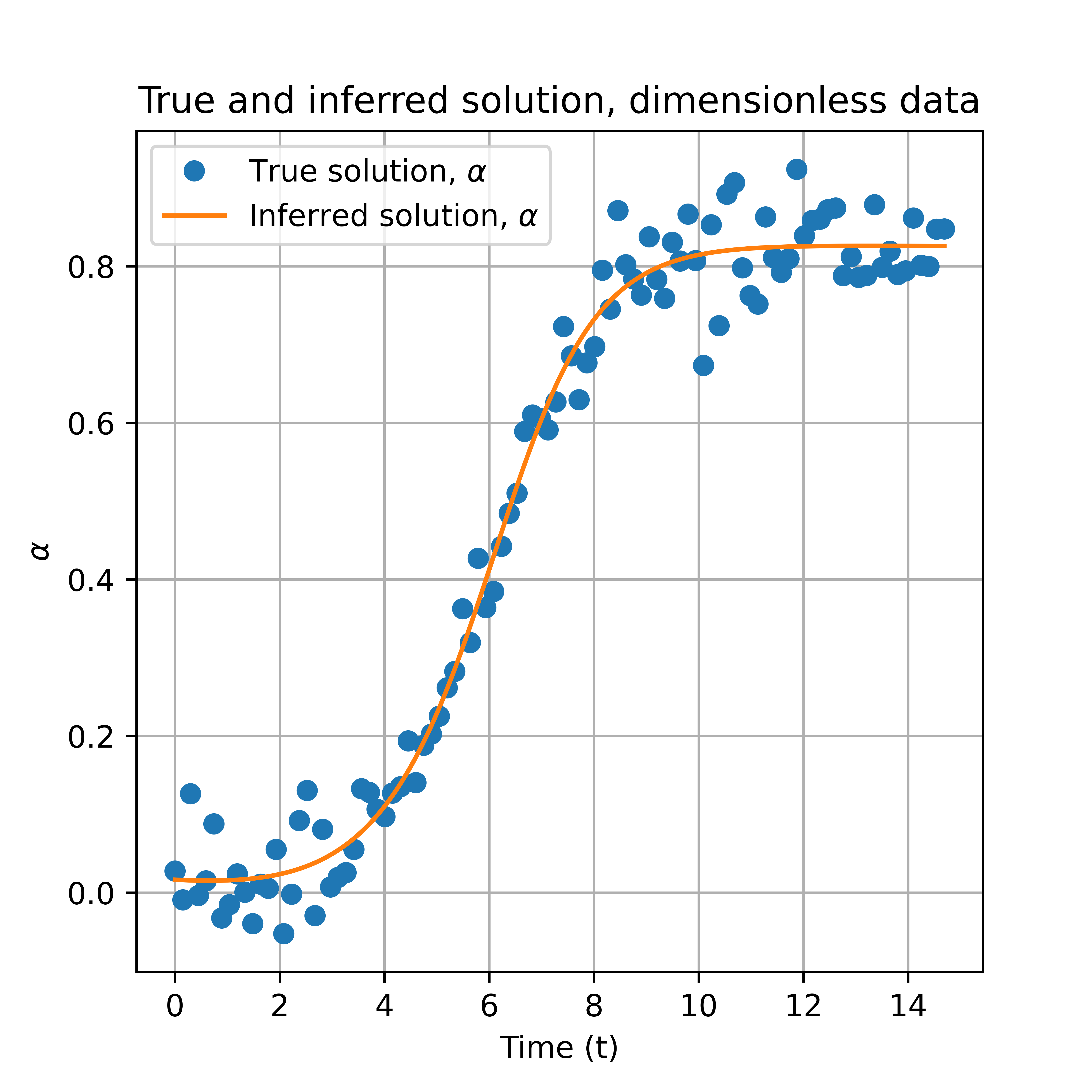}}
        % \caption{MSE: 0.002370}
        % \label{fig:linear}
    %--- Row 1 ---%
    \subfigure[]{\includegraphics[width=0.45\textwidth]{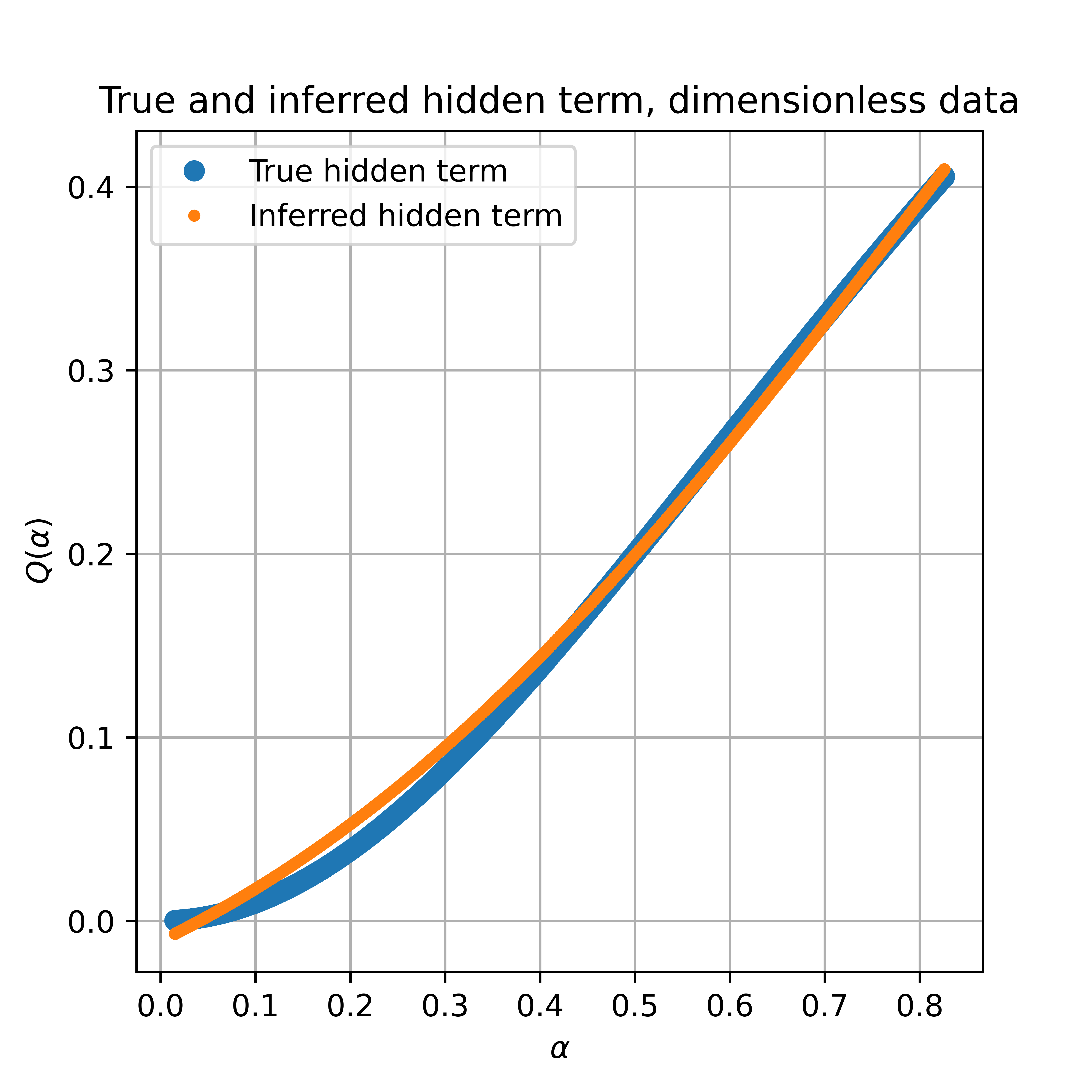}}

    \caption{Solution and hidden term fit for the logistic equation (noise level 0.10), for one set of parameters. (a) and (b) are the original equation, whereas (c) and (d) are the fits for the dimensionless equation.}
    \label{fig:logistic_growth_noise10}
\end{figure}

In Figure~\ref{fig:logistic_growth_noise5} and Figure~\ref{fig:logistic_growth_noise10} we show the noisy data for one set of parameters, as well as the learned solution and hidden term. Compared to Figure~\ref{fig:logistic_growth} the data is visibly noisy. However, we see that the UPINN performs well in recovering the solutions and the hidden terms. Table~\ref{tab:logg_mse_noisy} compares the MSE of the hidden terms for both noisy regimes. As noise level increases, the MSE of the recovered hidden terms increases, but for each noise level, nondimensionalizing the equation yields a smaller hidden term MSE by several orders of magnitude. The lowest MSE per regime is bolded. Finally, in Table~\ref{tab:noisy5} (noise level 0.05) and Table~\ref{tab:noisy10} (noise level 0.1) we compare the discovered equations for both the original and nondimensionalized form. For the nondimensionalized data, the ground truth equation is always among the expressions returned by PySR, but for the original data this is not the case. Hence, we conclude that our framework (compared to using symbolic regression on the data as is) is able to successfully recover a symbolic expression for the hidden term even in cases of significantly noisy data.

\begin{table}[htbp]
\centering
\begin{tabular}{ccccc}
\toprule
& \multicolumn{2}{c}{Noise level $\epsilon=0.05$} & \multicolumn{2}{c}{Noise level $\epsilon=0.1$} \\
\cmidrule(lr){2-3} \cmidrule(lr){4-5}
param set & MSE (orig.) & MSE (nondim.) & MSE (orig.) & MSE (nondim.) \\
\midrule
1 & $6.16 \times 10^{-4}$ & \boldmath$1.67 \times 10^{-5}$ & $2.59 \times 10^{-3}$ & \boldmath$3.48 \times 10^{-5}$ \\
2 & $6.10 \times 10^{-3}$ & \boldmath$4.57 \times 10^{-5}$ & $2.00 \times 10^{-2}$ & \boldmath$8.26 \times 10^{-5}$ \\
3 & $4.49 \times 10^{-4}$ & \boldmath$1.38 \times 10^{-6}$ & $1.64 \times 10^{-3}$ & \boldmath$6.95 \times 10^{-6}$ \\
4 & $2.75 \times 10^{-4}$ & \boldmath$2.79 \times 10^{-6}$ & $1.07 \times 10^{-3}$ & \boldmath$1.37 \times 10^{-5}$ \\
5 & $5.20 \times 10^{-2}$ & \boldmath$1.92 \times 10^{-4}$ & $1.04 \times 10^{-1}$ & \boldmath$4.47 \times 10^{-4}$ \\
6 & $2.33 \times 10^{-5}$ & \boldmath$2.21 \times 10^{-7}$ & $5.79 \times 10^{-5}$ & \boldmath$1.82 \times 10^{-6}$ \\
7 & $3.41 \times 10^{-3}$ & \boldmath$6.65 \times 10^{-6}$ & $8.38 \times 10^{-3}$ & \boldmath$2.40 \times 10^{-5}$ \\
8 & $2.27 \times 10^{-3}$ & \boldmath$1.19 \times 10^{-5}$ & $7.38 \times 10^{-3}$ & \boldmath$4.33 \times 10^{-5}$ \\
9 & $1.78 \times 10^{-2}$ & \boldmath$5.79 \times 10^{-5}$ & $8.54 \times 10^{-2}$ & \boldmath$2.23 \times 10^{-4}$ \\
10 & $8.83 \times 10^{-4}$ & \boldmath$1.33 \times 10^{-5}$ & $4.55 \times 10^{-3}$ & \boldmath$2.13 \times 10^{-5}$ \\
\bottomrule
\end{tabular}
\vspace{1em}
\caption{Parameter values and MSE for original and nondimensionalized fits of the unknown term for the logistic growth equation, with different noise levels. The parameter sets follow the rows in Table~\ref{tab:logg_mse}.}
\label{tab:logg_mse_noisy}
\end{table}

\begin{table}[htbp]
\centering
\begin{tabular}{rcl}
\toprule
Complexity & Loss & Discovered equation  (ground truth: $AN^2/(B^2+N^2)$)\\
\midrule
1 & $19.29$   & $4.1256075$ \\
2 & $13.32$      & $\sqrt{N}$ \\
3 & $2.84$   & $N \cdot 0.55867$ \\
5 & $1.87$   & $(N \cdot 0.03423748) \cdot A$ \\
6 & $1.03$   & $N \cdot (\text{inv}(B) \cdot 7.607662)$ \\
7 & $0.89$  & $((B \cdot -0.04339262) + 1.1776354) \cdot N$ \\
8 & $0.23$  & $N \cdot ((A \cdot 0.46449503) \cdot \text{inv}(B))$ \\
10 & $0.19$  & $(((N \cdot 0.4644951) \cdot \text{inv}(B)) \cdot A) + -0.2118686$ \\
\midrule
Complexity & Loss & Discovered equation (ground truth: $\alpha^2/(\alpha^2+1)$)\\
\midrule
1 & $6.65 \times 10^{-2}$  & $0.2541925$ \\
2 & $5.67 \times 10^{-2}$  & $\sin(\alpha)$ \\
3 & $8.99 \times 10^{-4}$ & $\alpha \cdot 0.46227956$ \\
5 & $5.47 \times 10^{-4}$ & $(\alpha \cdot 0.48866192) + -0.027856393$ \\
6 & $4.86 \times 10^{-4}$ & $\sin((\alpha \cdot 0.5367224) + -0.03966328)$ \\
7 & $3.48 \times10^{-5}$ & \boldmath$\alpha \cdot \text{inv}(\alpha + \text{inv}(\alpha))$ \\
\bottomrule
\end{tabular}
\vspace{1em}
\caption{Symbolic expressions returned by PySR for the extended logistic growth model (top: original data, and bottom: nondimensionalized data, 10 ODE solves for each). The data is slightly noisy, noise level $\epsilon=0.05$. \textit{inv} represents the function $1/x$. The true discovered equation is bolded, and is algebraically equivalent to the true unknown term, $\alpha^2/(1+\alpha^2)$.}
\label{tab:noisy5}
\end{table}

\begin{table}[htbp]
\centering
\begin{tabular}{rcl}
\toprule
Complexity & Loss & Discovered equation  (ground truth: $AN^2/(B^2+N^2)$)\\
\midrule
1 & $19.72$   & $4.171902$ \\
2 & $13.76$   & $\sqrt{N}$ \\
3 & $2.94$    & $N \cdot 0.5624422$ \\
5 & $1.96$    & $N \cdot (A \cdot 0.034462262)$ \\
6 & $1.43$    & $N \cdot \cos(\log(\log(B)))$ \\
7 & $0.89$    & $N \cdot ((B \cdot -0.04417722) + 1.1924701)$ \\
8 & $0.74$    & $(N + -1.2374399) \cdot \cos(B \cdot 0.062140066)$ \\
\midrule
Complexity & Loss & Discovered equation (ground truth: $\alpha^2/(\alpha^2+1)$)\\
\midrule
1 & $6.82 \times 10^{-2}$  & $0.25714397$ \\
2 & $5.56 \times 10^{-2}$  & $\sin(\alpha)$ \\
3 & $8.95 \times 10^{-4}$  & $\alpha \cdot 0.46581584$ \\
5 & $5.26 \times 10^{-4}$  & $(\alpha + -0.05793306) \cdot 0.49274248$ \\
6 & $4.79 \times 10^{-4}$  & $\sin((\alpha \cdot 0.5423931) + -0.04076865)$ \\
7 & \boldmath$8.98 \times 10^{-5}$  & \boldmath$\alpha \cdot \text{inv}(\alpha + \text{inv}(\alpha))$ \\
9 & $6.08 \times 10^{-5}$  & $\alpha \cdot \sin(\exp(\cos(\log(\alpha)) + -1.6348561))$ \\
\bottomrule
\end{tabular}
\vspace{1em}
\caption{Symbolic expressions returned by PySR for the extended logistic growth model (top: original data, and bottom: nondimensionalized data, 10 ODE solves for each). The data is more noisy, noise level $\epsilon=0.10$. \textit{inv} represents the function $1/x$. The true equation is bolded when it is discovered.}
\label{tab:noisy10}
\end{table}

\subsubsection{Rotating bead ODE results}

\begin{figure}[h]
    \centering
     \subfigure[]{\includegraphics[width=0.45\textwidth]{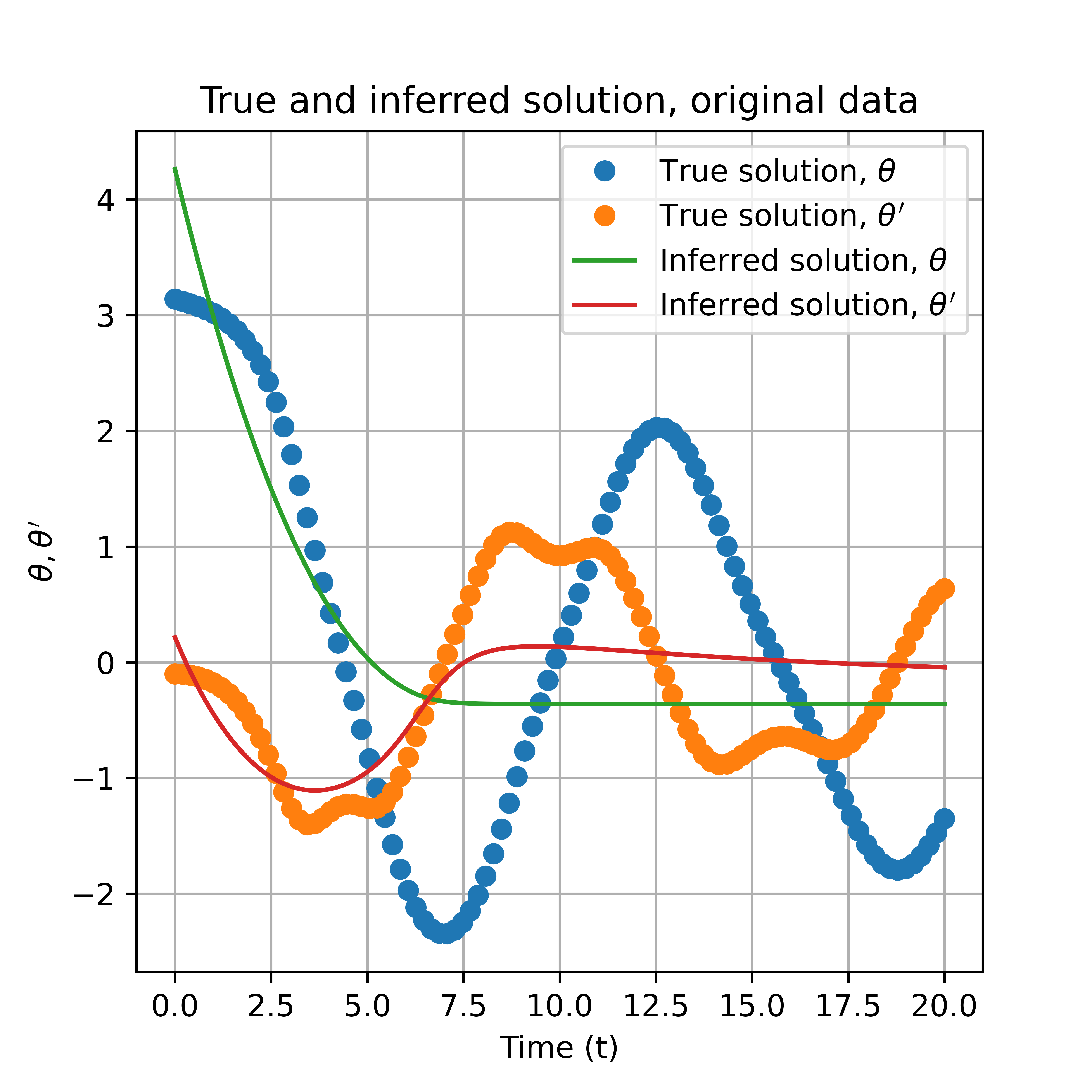}}
    \subfigure[]{\includegraphics[width=0.45\textwidth]{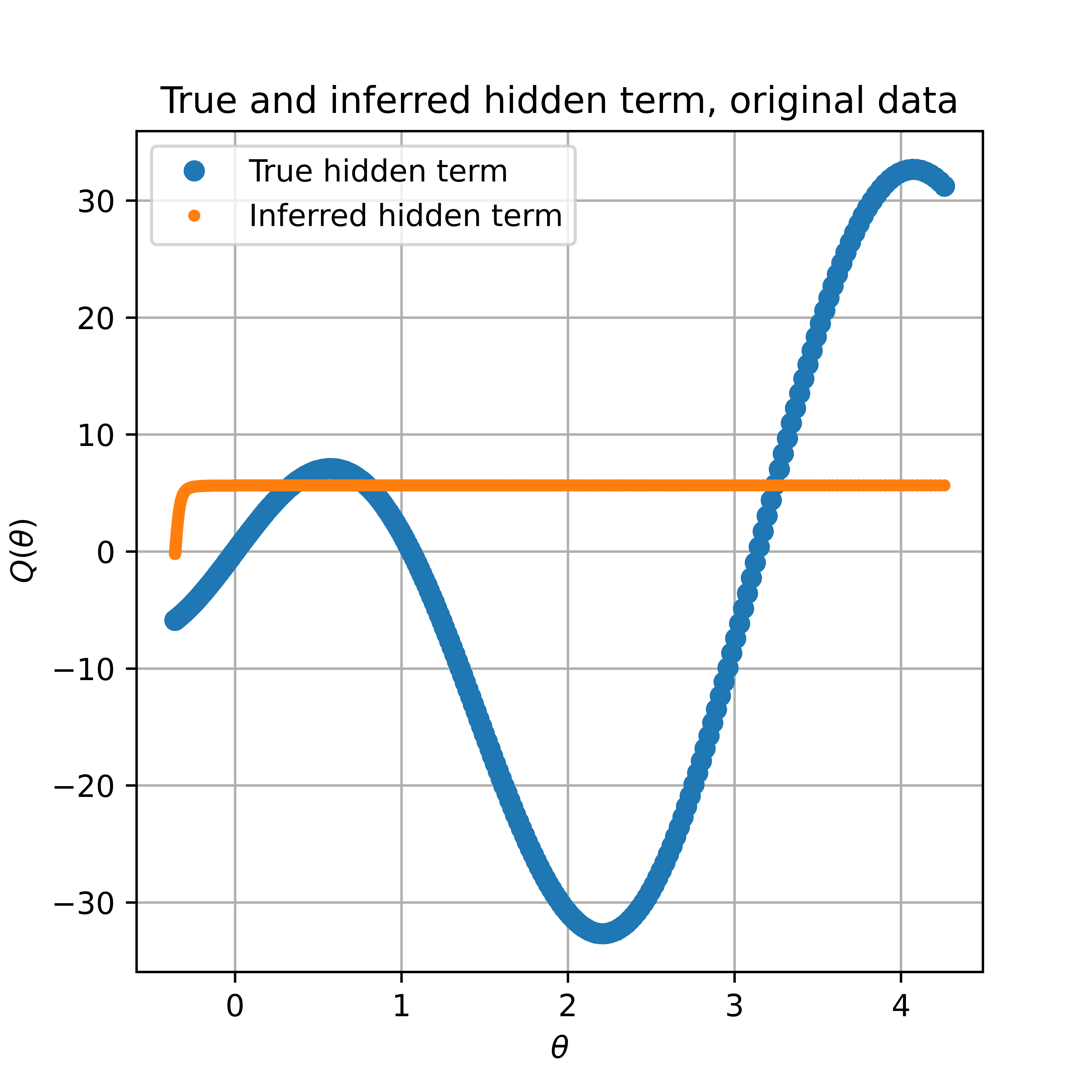}}
    \subfigure[]{\includegraphics[width=0.45\textwidth]{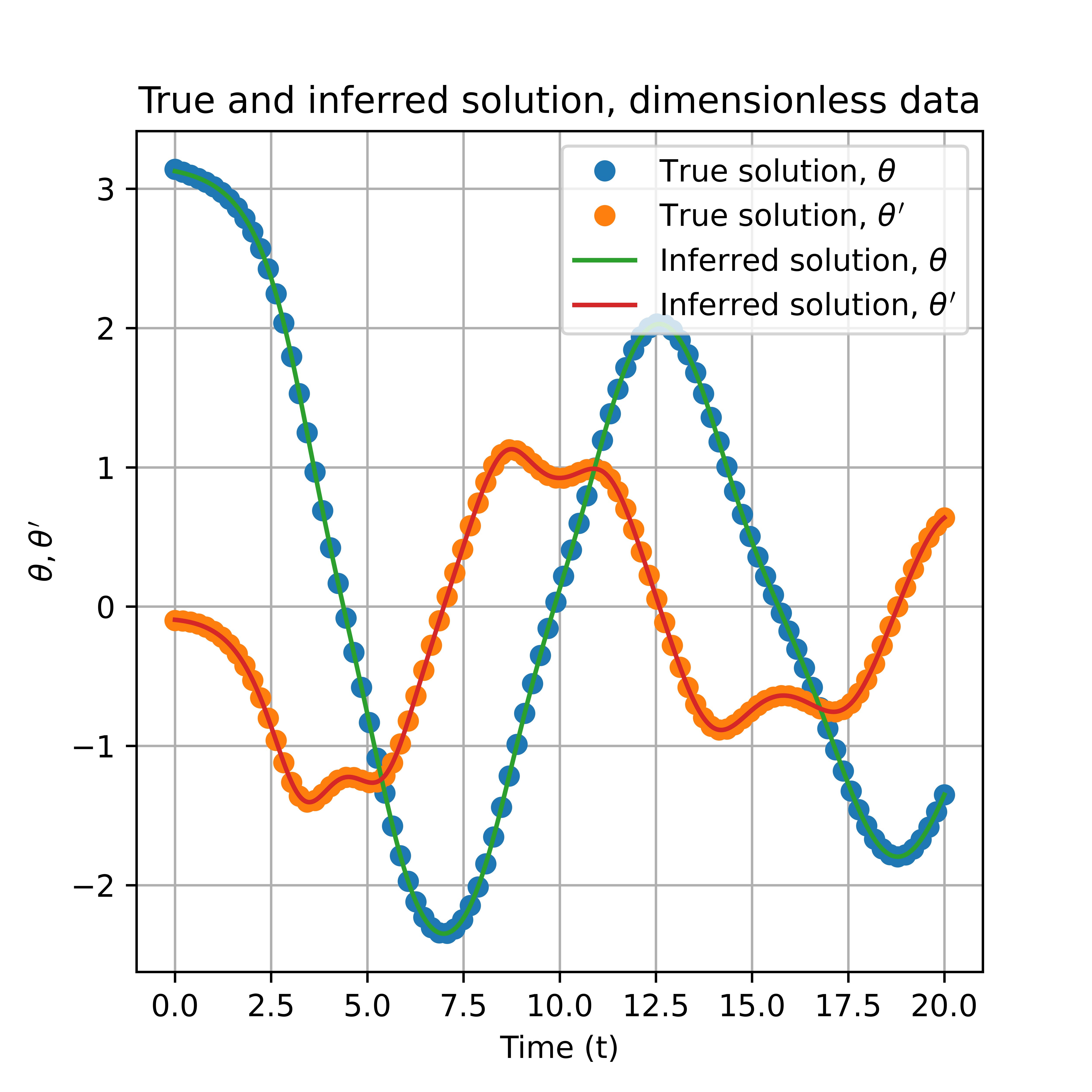}}
    \subfigure[]{\includegraphics[width=0.45\textwidth]{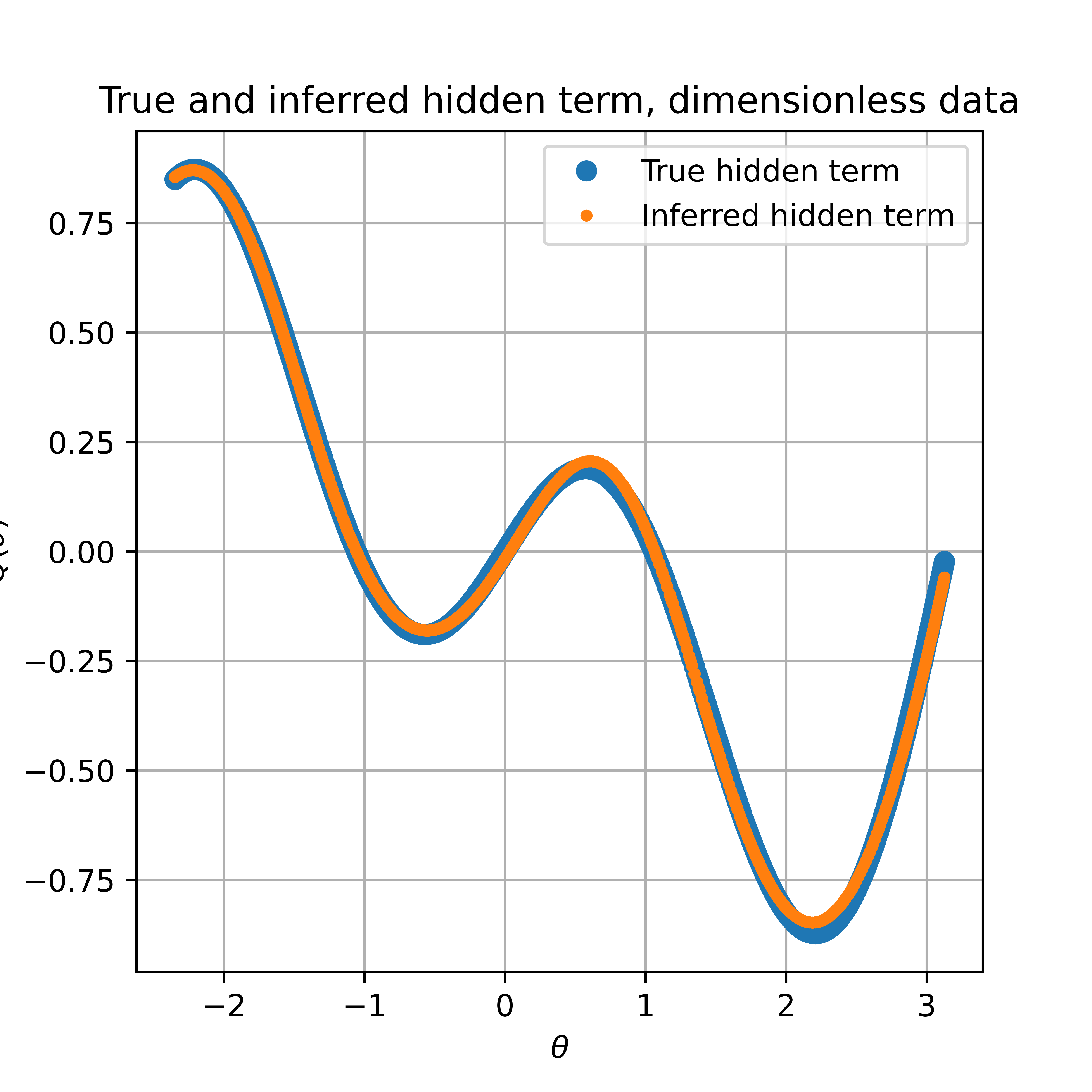}}

    \caption{The recovered UPINN fit for both the original and dimensionless rotating bead differential equations. (a) and (b) are the original equation, whereas (c) and (d) are the fits for the dimensionless equation.}
    \label{fig:bead}
\end{figure}

For the rotating bead model, we obtain observations of the unknown term using UPINNs and then we give it as input to PySR and recover the true unknown term, in both the nondimensionalized and original equation. Table~\ref{tab:parameters_bead} show the parameters of each ODE solve and the MSE in recovering the unknown term for the original and dimensionless equation. The MSE tends to be quite large ($>10$) for the original equation. Figure~\ref{fig:bead} shows the recovered unknown term for a single ODE solve both before and after nondimensionalization. We note that the nondimensionalization we chose earlier for this problem does not change the scale of the data at all. Despite this, the unknown term changed, and the solution is much more difficult for the UPINN to recover from the original data. This shows that nondimensionalization, aside from potentially changing the scale of the data, could significantly change the difficulty of recovering the unknown term. For this reason, PySR does not perform well in recovering the unknown term from the UPINN output. Table~\ref{tab:rotating_bead_nondim} shows the result of PySR attempting to recover the true equation from the original data and for the dimensionless equation. PySR is able to recover the dimensionless equation, but not the original one. In fact, there are no equations that fit the unknown term well for the original equation.

\begin{table}[htbp]
\centering
\begin{tabular}{cccccccc}
\toprule
$b$ & $r$ & $m$ & $\varepsilon$ & MSE (orig.) & MSE (nondim.) \\
\midrule
2.5707 & 19.9604 & 1.8755 & 0.4910 & $1.30 \times 10^{2}$ & $2.04 \times 10^{-4}$ \\
2.6037 & 78.2176 & 1.5529 & 0.1253 & $1.45 \times 10^{3}$ & $1.63 \times 10^{-3}$ \\
2.0020 & 95.8558 & 1.7156 & 0.1022 & $3.24 \times 10^{3}$ & $3.90 \times 10^{-3}$ \\
2.1224 & 68.6628 & 1.7405 & 0.1427 & $1.04 \times 10^{3}$ & $7.95 \times 10^{-2}$ \\
2.7653 & 50.8052 & 2.5457 & 0.1929 & $1.56 \times 10^{3}$ & $7.65 \times 10^{-3}$ \\
1.7376 & 37.1237 & 1.1508 & 0.2640 & $1.37 \times 10^{2}$ & $9.01 \times 10^{-4}$ \\
2.5775 & 93.3809 & 1.7944 & 0.1049 & $2.85 \times 10^{3}$ & $3.59 \times 10^{-4}$ \\
1.8723 & 32.3668 & 2.7383 & 0.3028 & $7.31 \times 10^{2}$ & $5.61 \times 10^{-4}$ \\
2.4092 & 80.4126 & 2.4085 & 0.1219 & $3.51 \times 10^{3}$ & $7.87 \times 10^{-2}$ \\
2.8186 & 22.6604 & 1.8843 & 0.4325 & $1.51 \times 10^{2}$ & $2.89 \times 10^{-5}$ \\
\bottomrule
\end{tabular}
\vspace{1em}
\caption{Parameter values and MSE for original and nondimensionalized fits for the rotating bead equation. Note that $g=9.8$ and $w=1.0$ are fixed across all runs and omitted from the table.}
\label{tab:parameters_bead}
\end{table}

\begin{table}[htbp]
\centering
\begin{tabular}{rcl}
\toprule
Complexity & Loss & Discovered equation: (ground truth: $mr \omega^2 \sin \theta \cos \theta  - m g \sin \theta$)  \\
\midrule
1 & 548.12   & $3.4967551$ \\
2 & 545.76   & $\log(r)$ \\
3 & 324.75  & $\theta \cdot -9.199068$ \\
4 & 204.39  & $\exp(\theta \cdot -1.6174623)$ \\
6 & 179.17  & $m \cdot \exp(\theta \cdot -1.3801409)$ \\
7 & 146.54  & $((\theta + -3.6895895) \cdot \theta) \cdot 3.4955583$ \\
8 & 144.21  & $\exp((\theta \cdot -0.78455627) + 2.3025424) + (-12.418046)$ \\
9 & 136.51  & $(\theta \cdot ((\theta + -6.8664646) + \theta)) \cdot m$ \\
\midrule
Complexity & Loss & Discovered equation (ground truth: $-\varepsilon \sin \theta + \sin \theta \cos \theta$)\\
\midrule
1 & $1.61\times 10^{-1}$ & $-0.030887194$ \\
3 & $6.90\times 10^{-2}$ & $\theta \cdot -0.16079584$ \\
4 & $6.89\times 10^{-2}$ & $\sin(\theta \cdot -0.16550536)$ \\
5 & $5.52\times 10^{-2}$ & $\sin(\theta) \cdot \cos(\theta)$ \\
6 & $5.21\times 10^{-2}$ & $\sin(\theta \cdot -2.0650928) \cdot -0.46723858$ \\
7 & $2.56\times 10^{-2}$ & $(\cos(\theta) -0.24893238) \cdot \sin(\theta)$ \\
8 & $2.25\times 10^{-2}$ & $\sin(\theta) \cdot \sin(\cos(\theta) -0.31793493)$ \\
\textbf{9} & \boldmath$1.73\times 10^{-2}$ & $\boldsymbol{\sin(\theta) \cdot (\cos(\theta) + (\varepsilon \cdot -1.0024753))}$ \\
10 & $1.50\times 10^{-2}$ & $((\varepsilon \cdot -1.1765212) + \cos(\theta)) \cdot \sin(\sin(\theta))$ \\
\bottomrule
\end{tabular}
\vspace{1em}
\caption{Symbolic expressions returned by PySR for the rotating bead problem (top: original, bottom: nondimensionalized, 10 ODE solves). The correct expression among the candidates is bolded.}
\label{tab:rotating_bead_nondim}
\end{table}

\subsection{Extended Lotka-Volterra Model}

\begin{figure}[h]
    \centering
     \subfigure[]{\includegraphics[width=0.45\textwidth]{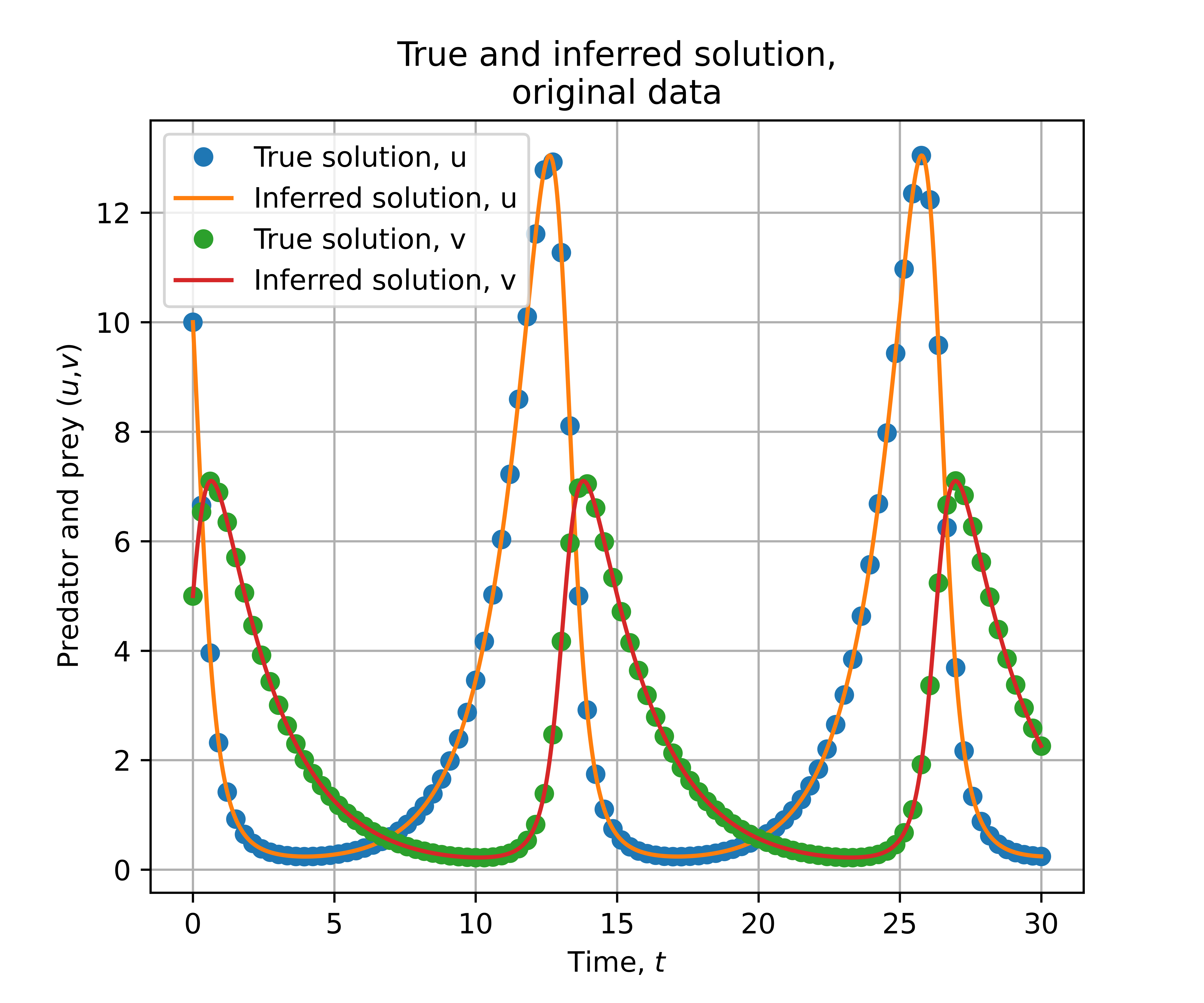}}
        % \caption{MSE: 0.002370}
        % \label{fig:linear}
    %--- Row 1 ---%
    \subfigure[]{\includegraphics[width=0.45\textwidth]{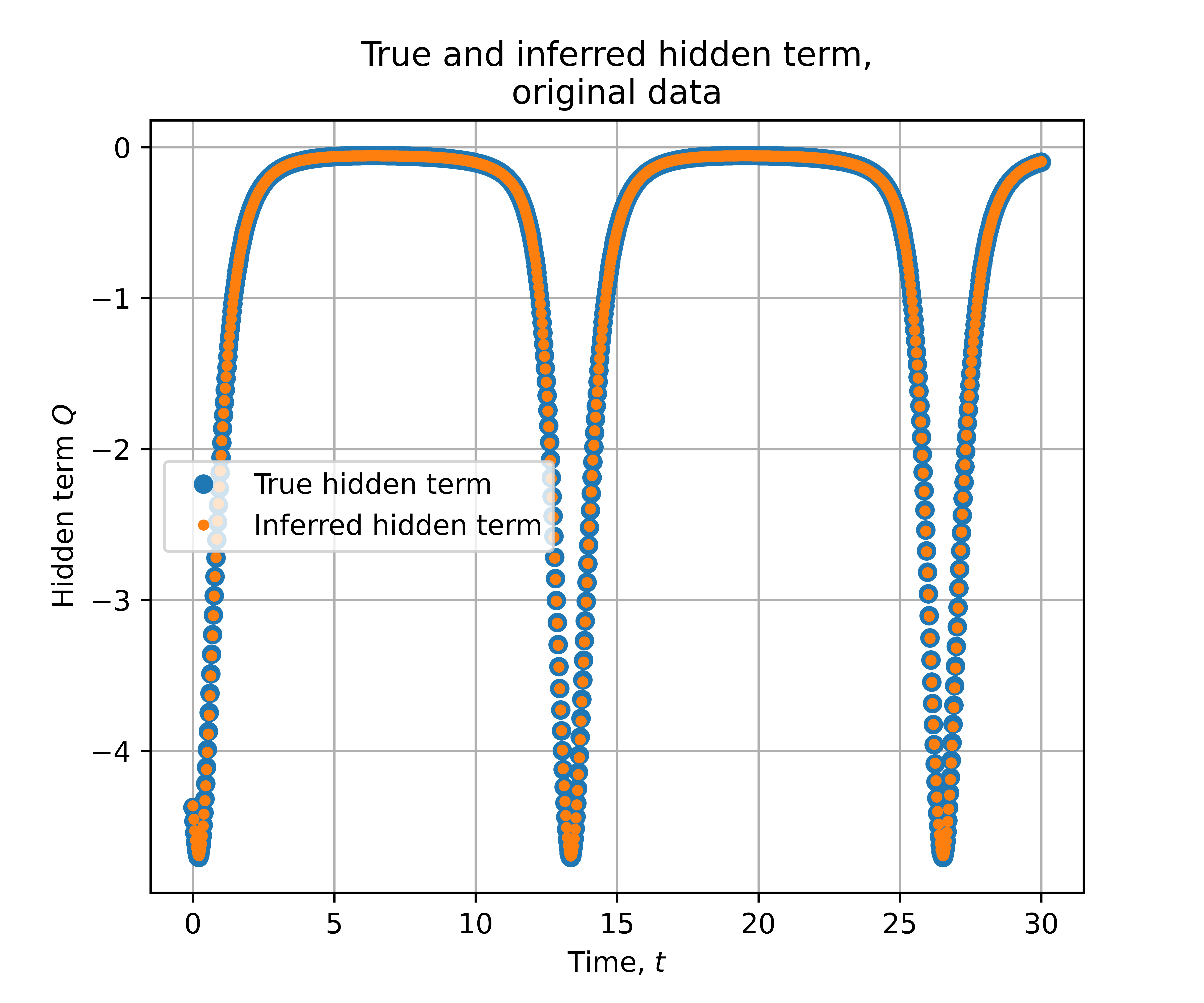}}
    \subfigure[]{\includegraphics[width=0.45\textwidth]{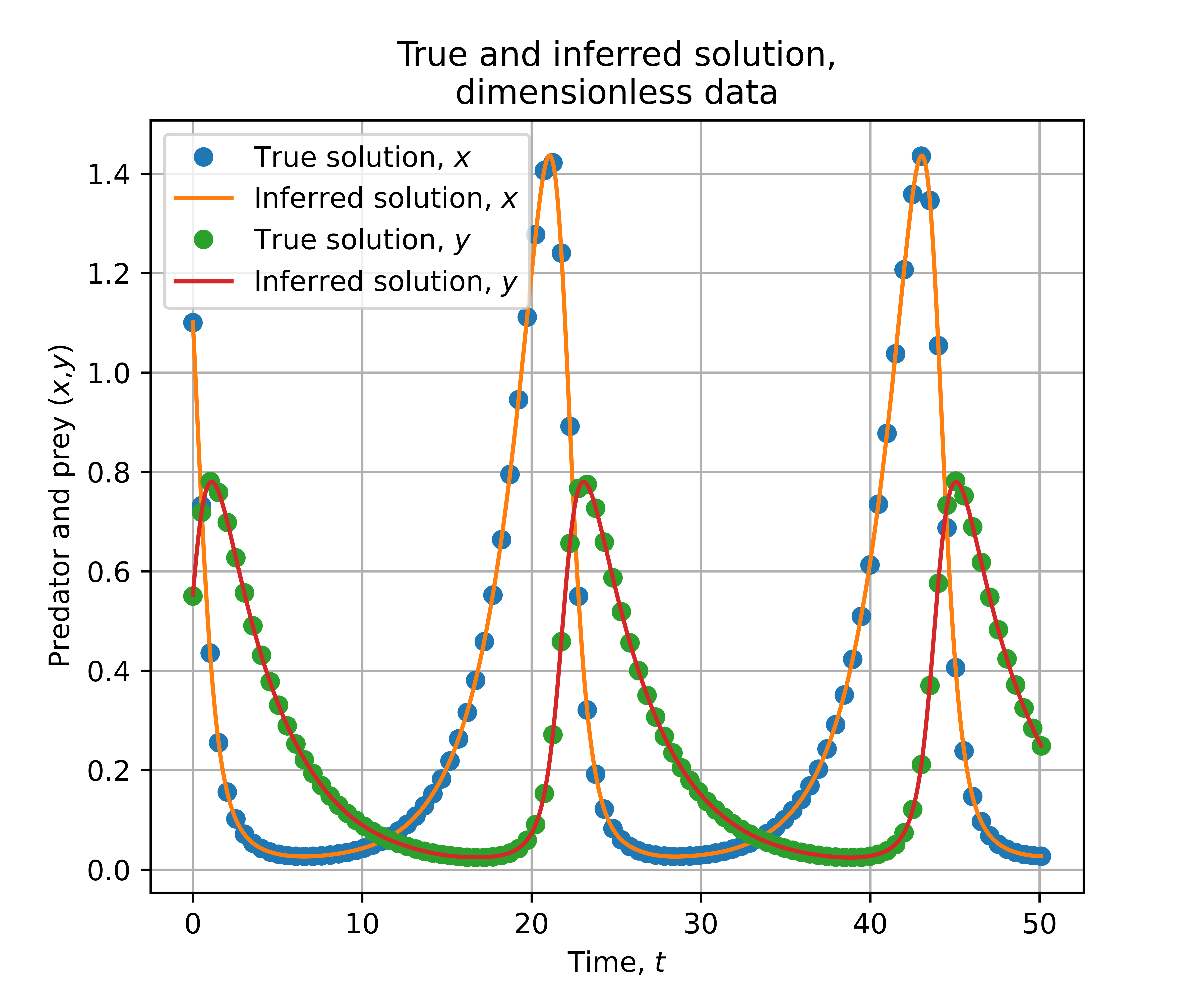}}
        % \caption{MSE: 0.002370}
        % \label{fig:linear}
    %--- Row 1 ---%
    \subfigure[]{\includegraphics[width=0.45\textwidth]{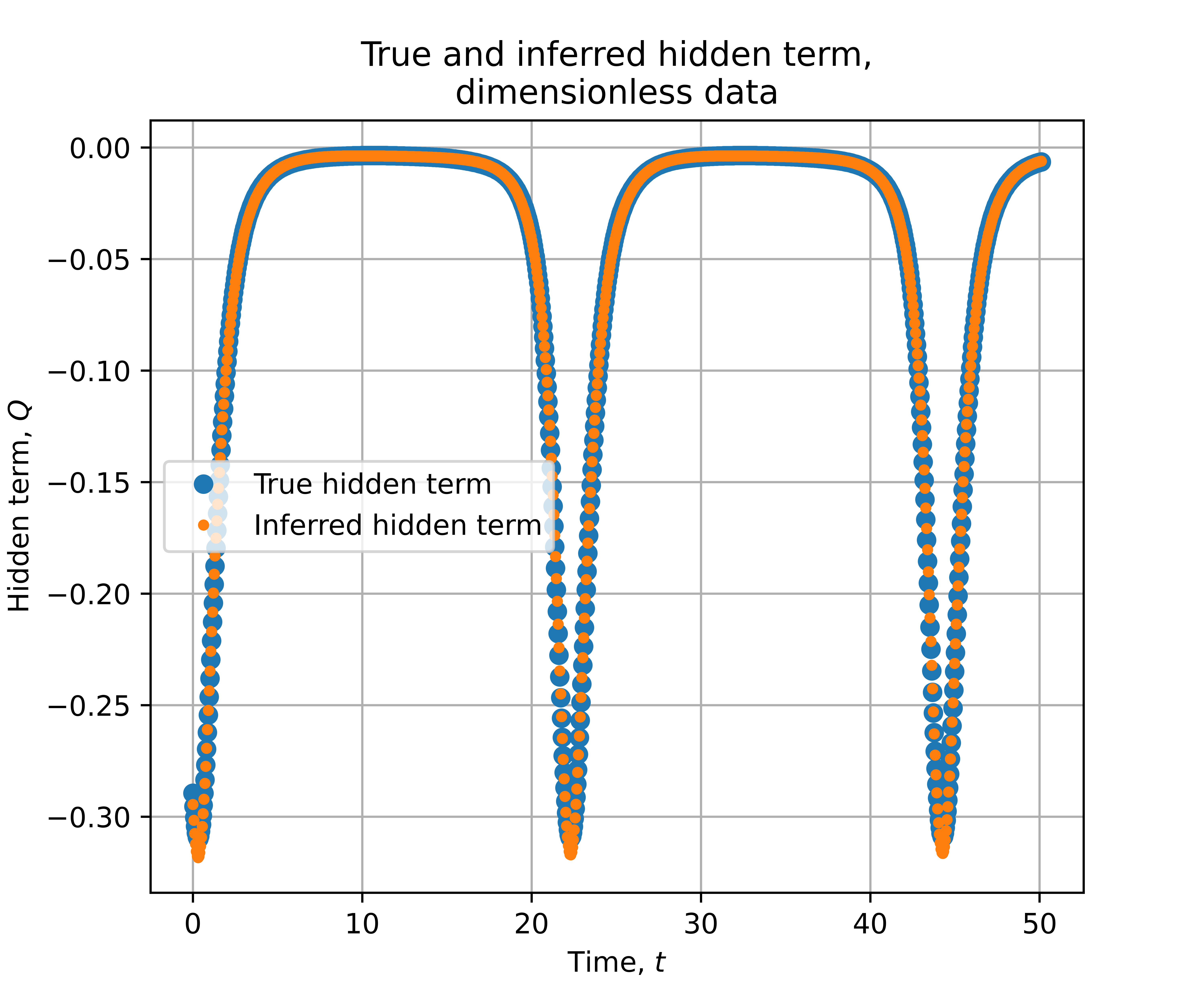}}

    \caption{Solution and hidden term fit for the Lotka-Volterra equation (noiseless data), for one set of parameters. (a) and (b) are the original equation, whereas (c) and (d) are the fits for the dimensionless equation. The hidden term is plotted as a function of time for better visualization, although it was learned as a function of $Q(u,v)$ and $Q(x,y)$ in the regime of the original data and nondimensionalized data respectively.}
    \label{fig:lotka_volterra_soln_ht}
\end{figure}

Following data generation for the Lotka-Volterra extended model as described in Section~\ref{sec:lotka_volterra}, we evaluate the example in the same way as the previous two. Figure~\ref{fig:lotka_volterra_soln_ht} shows the learned surrogate solution and hidden term for one ODE solve, in both its original form and the nondimensionalized form. Visually we can observe that the hidden term can be inferred well in both cases, despite the periodicity and range of the data. Table~\ref{tab:lv_mse} shows the mean squared error in recovering the hidden term for each set of parameters, for both the original data and its nondimensionalized form. We can see that in this scenario, the UPINN occasionally performs better in recovering the original data, occasionally in by several orders of magnitude. However, in general the MSE stays below $10^{-2}$ for both cases. Interestingly, this may be due to the rescaling of the data and hidden term to a smaller range than the MLP may easily approximate (as can be seen in Figure~\ref{fig:lotka_volterra_soln_ht}). The ability of the UPINN to recover the hidden term with high accuracy is contrasted with the ability of PySR to recover the true equation (Table~\ref{tab:lotka_volterra_pysr}). PySR does not recover the true interaction term in either case. This is potentially due to insufficient information within the dataset to differentiate between the ground truth data and a similar expression. Evidence for this can be seen in the lower section of Table~\ref{tab:lotka_volterra_pysr}, as PySR has correctly identified the hidden term as being proportional to $xy$, but even a formula of the form $Cxy$ for some constant $C$ (4th row of the table) has a loss of $3\times 10^{-3}$. A more complex expression (introducing trigonometric functions, as can be seen in later rows) is hardly able to improve upon this performance. Hence, the results for this example are somewhat mixed, but we see that our framework improves the overall loss of PySR's discovered expressions, although the true expression is not identified in the case of original data, nor in the case of nondimensionalized data.

\begin{table}[htbp]
\centering
\begin{tabular}{cccccccc}
\toprule
$\alpha$ & $\beta$ & $\gamma$ & $\delta$ & $\theta$ & $h$ & MSE (orig.) & MSE (nondim.) \\
\midrule
0.6915 & 0.3488 & 0.4626 & 0.2571 & 1.6700 & 9.0889  & $2.33\times 10^{-5}$ & \boldmath$3.28\times 10^{-6}$ \\
0.7765 & 0.4207 & 0.7749 & 0.2752 & 1.0367 & 12.5149 & \boldmath$3.02\times 10^{-4}$ & $4.50\times 10^{-4}$ \\
1.1835 & 0.3851 & 0.4222 & 0.2122 & 1.2546 & 5.2065  & $6.52\times 10^{-5}$ & \boldmath$5.92\times 10^{-6}$ \\
1.2728 & 0.4531 & 0.4189 & 0.2231 & 0.6131 & 10.5324 & \boldmath$4.65\times 10^{-4}$ & $3.02\times 10^{-3}$ \\
1.4331 & 0.3606 & 0.4383 & 0.2577 & 0.9753 & 13.5215 & \boldmath$2.87\times 10^{-4}$ & $2.05\times 10^{-3}$ \\
1.3691 & 0.2745 & 0.6813 & 0.1288 & 1.5564 & 15.5687 & $9.61\times 10^{-7}$ & \boldmath$2.81\times 10^{-9}$ \\
0.7188 & 0.4699 & 0.4653 & 0.2819 & 0.5897 & 7.7643  & \boldmath$2.33\times 10^{-5}$ & $3.02\times 10^{-3}$ \\
0.5474 & 0.3700 & 0.5568 & 0.2067 & 0.5650 & 13.4215 & \boldmath$1.98\times 10^{-5}$ & $3.56\times 10^{-3}$ \\
0.8297 & 0.3012 & 0.2671 & 0.2214 & 1.3489 & 5.1015  & \boldmath$2.02\times 10^{-5}$ & $1.30\times 10^{-2}$ \\
1.1174 & 0.4648 & 0.6743 & 0.2984 & 1.9382 & 16.8795 & $5.12\times 10^{-2}$ & \boldmath$3.16\times 10^{-4}$ \\
\bottomrule
\end{tabular}
\vspace{1em}
\caption{Parameter values and MSE for original and nondimensionalized fits for the Lotka--Volterra model. The smaller MSE in each row is shown in bold.}
\label{tab:lv_mse}
\end{table}

\begin{table}[htbp]
\centering
\begin{tabular}{rcl}
\toprule
Complexity & Loss & Discovered equation (ground truth: $-\theta u v /(h+u)$) \\
\midrule
1 & $1.3961$             & $-0.81001276$ \\
3 & $8.24\times 10^{-1}$ & $u \cdot -0.22557844$ \\
5 & $4.14\times 10^{-1}$ & $(v \cdot u) \cdot -0.06304521$ \\
6 & $4.00\times 10^{-1}$ & $(\sqrt{u} \cdot -0.17813776) \cdot v$ \\
7 & $3.67\times 10^{-1}$ & $((\theta + u) \cdot -0.05459503) \cdot v$ \\
8 & $3.34\times 10^{-1}$ & $(\sqrt{\theta \cdot u} \cdot -0.1591276) \cdot v$ \\
9 & $3.07\times 10^{-1}$ & $(\sin(\theta) \cdot (v \cdot \sqrt{u})) \cdot -0.19914308$ \\
\midrule
Complexity & Loss & Discovered equation (ground truth: $-xy/(1+x)$)\\
\midrule
1  & $1.34\times 10^{-2}$  & $-0.061528556$ \\
3  & $5.90\times 10^{-3}$  & $x \cdot -0.19865212$ \\
4  & $5.90\times 10^{-3}$  & $\sin(x \cdot -0.20126656)$ \\
5  & $2.80\times 10^{-3}$  & $(y \cdot x) \cdot -0.40584636$ \\
6  & $2.27\times 10^{-3}$  & $\sin(y \cdot x) \cdot -0.5389494$ \\
7  & $2.14\times 10^{-3}$  & $(\sin(y) \cdot \sin(x)) \cdot -0.69302034$ \\
8  & $2.06\times 10^{-3}$  & $(\sin(x) \cdot \sin(\sin(y))) \cdot -0.7624622$ \\
9  & $1.97\times 10^{-3}$  & $\sin(\sin(x \cdot -0.7277368)) \cdot \sin(\sin(y))$ \\
10 & $1.95\times 10^{-3}$  & $\sin(\sin(\sin(y)) \cdot \sin(\sin(x \cdot -0.75338626)))$ \\
\bottomrule
\end{tabular}
\vspace{1em}
\caption{Symbolic expressions returned by PySR for the Lotka-Volterra differential equation (top: original data, bottom: nondimensionalized data).}
\label{tab:lotka_volterra_pysr}
\end{table}

\section{Conclusion and Discussion}

Symbolic regression can be an invaluable tool in engineering and applied mathematics, allowing practitioners to use machine learning to automatically infer the equations that relate variables within a dataset. To enhance the accuracy and decrease the runtime of symbolic regression, we demonstrate the integration of dimensional analysis with the PySR symbolic regression pipeline. However, the approach can be used on any symbolic regression algorithm. This integration also allows one to discover equations in cases of partial knowledge, via Universal Physics-Informed Neural Networks.
The framework combines symbolic regression with dimensional analysis, utilizing both the Buckingham $\Pi$ theorem and Ipsen's method. This approach effectively reduces the complexity of the search space and ensures physical consistency in the discovered equations. Our findings indicate that nondimensionalizing datasets prior to symbolic regression significantly enhances both accuracy and computational efficiency.

For algebraic equations, the application of the Buckingham $\Pi$ theorem notably decreased the number of independent variables, thus streamlining the symbolic regression process. This dimensionality reduction facilitated faster convergence, reduced the risk of overfitting, and improved accuracy, even in scenarios involving sparse datasets. We successfully recovered the symbolic expressions for several physics-based equations, demonstrating that this combined methodology can efficiently discover physically meaningful equations in complex systems with minimal computational resources.

In the context of differential equations, we expanded this framework by integrating dimensional analysis and Universal Physics-Informed Neural Networks (UPINNs) under a partial knowledge scenario. Ipsen’s method systematically reduced the number of unknown parameters by transforming the equations into their dimensionless forms and offered guidance on how to best nondimensionalize the partially known equation. The synergy between dimensional analysis and symbolic regression enhanced the capability of UPINNs to efficiently uncover unknown terms in differential equations, leading to faster convergence and a diminished reliance on extensive datasets. This is particularly beneficial in situations where data acquisition is limited or expensive. In comparison to other variable transformations, we highlight that the Buckingham $\Pi$ theorem performs reductions that are guaranteed to preserve the original relationship; arbitrary transformations of the variables may complicate the relationship between the variables, or may not hold sufficient information to recover the original relationship. A simple counterexample of this is the following dataset: $x_1 \sim \mathcal{N}(0,1), x_2 = -x_1, y =x_1x_2$. A transformation such as $a_1=x_1+x_2$ loses information, and will make it impossible to recover the correct relationship between $a_1$ and $y$ such that $\hat{y}=f_\theta(a_1)$ matches the dataset.

Additionally, while other works, such as those by Bakarji et al.~\cite{bakarji2022dimensionally}, apply symbolic regression and the Buckingham $\Pi$ theorem to scenarios like the rotating bead problem, our method introduces key distinctions. Bakarji et al. focus on identifying the most \textit{physically meaningful} dimensionless terms that best match the data given an optimal relationship between the terms. In contrast, our approach utilizes the Buckingham $\Pi$ theorem to uncover an unknown term in a partially known model, treating all nondimensionalizations as equivalent and favoring those that minimize the number of $\pi$ groups involved in the unknown term. This simplification aids UPINNs in learning fewer variable interactions, especially under conditions of sparse data availability. Initially applying UPINNs allows for the learning of potentially complex unknown terms with minimal inputs, followed by a powerful and versatile symbolic regression technique, such as PySR, to precisely identify the true unknown term without optimizing for the "most physically meaningful" dimensionless $\pi$ groups. Moreover, PySR could be substituted with other equally versatile symbolic regression algorithms like SymbolicGPT~\cite{valipour2021symbolicgpt} or AI Feynman~\cite{udrescu2020ai}.

Our results confirm that the combination of symbolic regression and dimensional analysis offers a robust methodology for automatically discovering governing equations in various domains, including physics. By leveraging the inherent scaling properties of physical laws, we can significantly simplify symbolic regression tasks, resulting in more interpretable and accurate models. In disciplines such as engineering and computational biology, the task of fitting a model to data can be greatly simplified and made more robust using an integrated framework such as the one we propose.

Several promising directions remain open for future work: enabling the relaxation of assumptions, real-world testing by constructing a thorough benchmark, and improving the robustness of the pipeline. The pipeline makes several assumptions: we assume that the inputs to the known and unknown terms are fully and exactly known; the dimensional matrix is assumed to be fully known; furthermore, we assume that there are no hidden or unobserved variables with additional dimensions. This complicates the application of the method to real-world data. Further investigation should be conducted to determine just by much these assumptions can be relaxed. If there are hidden or missing variables, it may be impossible to properly apply the Buckingham $\Pi$ theorem to nondimensionalize the equation, but the resulting data could potentially still be used to discover these hidden variables. For example, in the scenario where three variables collapse to one $\pi$ term after nondimensionalization, the presence of a variable can be noted if different sets of variable observations yield vastly different estimates for this singular $\pi$ term. Future work could potentially investigate handling the presence of unobserved (possibly dimensional) latent variables and inferring their dimensions. On the symbolic regression front, there is a need for a symbolic regression benchmark where real-world datasets are published together with their corresponding ground truth equations. Neither SRBench~\cite{srbench} nor its extension SRBench++~\cite{srbench++} contains such a dataset. Our method, and more generally any symbolic regression algorithm implemented within our framework, could then be tested on real-world data. However, many equations contain unobserved constants with dimensions (e.g., gravitational constant $G$). Hence, real-world testing could be done after extending our method to handle latent or hidden variables. We leave this investigation for future work. It may also be interesting to explore in depth the effect of different nondimensionalizations on symbolic regression performance. From preliminary experiments, we have noted that the resulting scale of the hidden term and data has a large effect on UPINN and PySR accuracy, but more investigation is needed. Finally, future work could look at decoupling the factors that affect the performance of UPINNs and the factors that affect the performance of symbolic regression algorithms, and thereby improving the pipeline so that it is more robust. Building on this proof-of-concept work, additional testing on a broad range of symbolic regression algorithms could also yield insights into the pipeline's robustness.

\section*{Acknowledgments}

This research was funded by the Natural Sciences and Engineering Research Council of Canada (NSERC), under the Discovery Grants Program, Grant No. RGPIN-2021-03472. 

\section*{CRediT Author Statement}
\begin{itemize}
    \item Joshveer Grewal: Symbolic regression analysis, writing original draft
    \item Diba Darooneh: Symbolic regression analysis, writing original draft
    \item Lena Podina:
    Symbolic regression analysis, UPINN simulations, formal analysis, writing original draft, review and editing
    \item Mohammad Kohandel: Project conceptualization, formal analysis, review and editing
\end{itemize}

\section*{Data and Code Availability}

The code used to generate the data will be made public on GitHub upon publication of the manuscript.

%\vspace{15mm}
\bibliographystyle{unsrtnat}
\bibliography{references}

@article{rackauckas2020universal,
  author = {C. Rackauckas and Y. Ma and J. Martensen},
  title = {{Universal Differential Equations for Scientific Machine Learning}},
  journal = {arXiv Prepr. arXiv:2001.04385},
  year = {2020}
}

@inproceedings{clark2022alternate,
  author = {J.P. Clark},
  title = {{An Alternate Means to Form Non-Dimensional Products in Dimensional Analysis}},
  booktitle = {ASME Turbo Expo: Power Land Sea Air},
  volume = {86021},
  pages = {V005T08A003},
  year = {2022},
  organization = {ASME}
}

@article{sonin2001physical,
  title = {{The Physical Basis of Dimensional Analysis}},
  author = {Sonin, A.A.},
  journal = {Department of Mechanical Engineering, MIT},
  pages = {1–36},
  year = {2001},
  url = {http://web.mit.edu/2.25/www/pdf/DA_unified.pdf}
}

@article{virgolin2022symbolic,
  author = {M. Virgolin, S.P. Pissis},
  title = {{Symbolic {Regression} is {NP}-Hard}},
  journal = {arXiv Prepr. arXiv:2207.01018},
  year = {2022}
}

@article{tenachi2023deep,
  author = {W. Tenachi, R. Ibata, F.I. Diakogiannis},
  title = {{Deep Symbolic Regression for Physics Guided by Units Constraints: Toward the Automated Discovery of Physical Laws}},
  journal = {Astrophys. J.},
  volume = {959},
  number = {2},
  pages = {99},
  year = {2023},
  publisher = {IOP Publ.}
}

@article{keren2023computational,
  author = {L.S. Keren, A. Liberzon, T. Lazebnik},
  title = {{A Computational Framework for Physics-Informed Symbolic Regression with Straightforward Integration of Domain Knowledge}},
  journal = {Sci. Rep.},
  volume = {13},
  number = {1},
  pages = {1249},
  year = {2023},
  publisher = {Nat. Publ. Group}
}

@article{udrescu2020ai,
  author = {S.-M. Udrescu, M. Tegmark},
  title = {{AI} {Feynman}: {A Physics-Inspired Method for Symbolic Regression}},
  journal = {Sci. Adv.},
  volume = {6},
  number = {16},
  pages = {eaay2631},
  year = {2020},
  publisher = {AAAS}
}

@book{barmeir2022fluidmechanics,
  author = {G. Bar-Meir},
  title = {{Basics of Fluid Mechanics}},
  publisher = {Potto Proj. Publ.},
  note = {doi:10.5281/zenodo.5521908},
  url = {https://doi.org/10.5281/zenodo.6462400},
  edition = {0.6.2},
  year = {2022},
  month = {Apr}
}

@book{Ipsens_Method,
  author = {F.M. White, H. Xue},
  title = {{Fluid Mechanics, 9th Ed.}},
  publisher = {McGraw-Hill},
  year = {2021}
}

@inproceedings{pmlr-v202-podina23a,
  author = {L. Podina, B. Eastman, M. Kohandel},
  title = {{Universal Physics-Informed Neural Networks: Symbolic Differential Operator Discovery with Sparse Data}},
  booktitle = {Proc. 40th Int. Conf. Mach. Learn.},
  volume = {202},
  pages = {27948--27956},
  year = {2023},
  month = {Jul},
  publisher = {PMLR},
  url = {https://proceedings.mlr.press/v202/podina23a.html}
}

@misc{cranmerInterpretableMachineLearning2023,
  author = {M. Cranmer},
  title = {{Interpretable Machine Learning for Science with {PySR} and {SymbolicRegression.jl}}},
  publisher = {arXiv},
  year = {2023},
  month = {May},
  note = {arXiv:2305.01582 [astro-ph, physics:physics]},
  url = {http://arxiv.org/abs/2305.01582},
  doi = {10.48550/arXiv.2305.01582}
}

@article{maslyaev2021partial,
  author = {M. Maslyaev, A. Hvatov, A.V. Kalyuzhnaya},
  title = {{Partial Differential Equations Discovery with {EPDE} Framework: Application for Real and Synthetic Data}},
  journal = {J. Comput. Sci.},
  pages = {101345},
  year = {2021},
  publisher = {Elsevier}
}

@inproceedings{augusto2000symbolic,
  title={Symbolic regression via genetic programming},
  author={Augusto, Douglas Adriano and Barbosa, Helio JC},
  booktitle={Proceedings. Vol. 1. Sixth Brazilian symposium on neural networks},
  pages={173--178},
  year={2000},
  organization={IEEE}
}

@inproceedings{mckay1995using,
  title={Using a tree structured genetic algorithm to perform symbolic regression},
  author={McKay, Ben and Willis, Mark J and Barton, Geoffrey W},
  booktitle={First international conference on genetic algorithms in engineering systems: innovations and applications},
  pages={487--492},
  year={1995},
  organization={IET}
}

@article{Quade_2016,
   title={Prediction of dynamical systems by symbolic regression},
   volume={94},
   ISSN={2470-0053},
   url={http://dx.doi.org/10.1103/PhysRevE.94.012214},
   DOI={10.1103/physreve.94.012214},
   number={1},
   journal={Physical Review E},
   publisher={American Physical Society (APS)},
   author={Quade, Markus and Abel, Markus and Shafi, Kamran and Niven, Robert K. and Noack, Bernd R.},
   year={2016},
   month=jul }

@article{valipour2021symbolicgpt,
  title={Symbolicgpt: A generative transformer model for symbolic regression},
  author={Valipour, Mojtaba and You, Bowen and Panju, Maysum and Ghodsi, Ali},
  journal={arXiv preprint arXiv:2106.14131},
  year={2021}
}

@article{ferro2019assessing,
  title={Assessing flow resistance law in vegetated channels by dimensional analysis and self-similarity},
  author={Ferro, Vito},
  journal={Flow Measurement and Instrumentation},
  volume={69},
  pages={101610},
  year={2019},
  publisher={Elsevier}
}

@article{muagurean2023three,
  title={Three-dimensional numerical model for heat transfer dynamic analysis for the building-soil interaction by applying the similarity theory},
  author={M{\u{a}}gurean, AM},
  journal={Thermal Science and Engineering Progress},
  volume={43},
  pages={102013},
  year={2023},
  publisher={Elsevier}
}

@article{schmidt2009distilling,
  title={Distilling free-form natural laws from experimental data},
  author={Schmidt, Michael and Lipson, Hod},
  journal={Science},
  volume={324},
  number={5923},
  pages={81--85},
  year={2009},
  publisher={American Association for the Advancement of Science}
}

@article{raissi2019physics,
  title={Physics-informed neural networks: A deep learning framework for solving forward and inverse problems involving nonlinear partial differential equations},
  author={Raissi, Maziar and Perdikaris, Paris and Karniadakis, George E},
  journal={Journal of Computational physics},
  volume={378},
  pages={686--707},
  year={2019},
  publisher={Elsevier}
}

@article{petersen2021deep,
  title={Deep Symbolic Regression: Recovering Mathematical Expressions from Data via Risk-Seeking Policy Gradients},
  author={Petersen, Bent and Pedersen, Morten and Hansen, Lars Kai},
  journal={Available on Papers with Code},
  year={2021}
}

@misc{li2024neuralguideddynamicsymbolicnetwork,
      title={A Neural-Guided Dynamic Symbolic Network for Exploring Mathematical Expressions from Data}, 
      author={Wenqiang Li and Weijun Li and Lina Yu and Min Wu and Linjun Sun and Jingyi Liu and Yanjie Li and Shu Wei and Yusong Deng and Meilan Hao},
      year={2024},
      eprint={2309.13705},
}

@article{kamienny2022end,
  title={End-to-end symbolic regression with transformers},
  author={Kamienny, Pierre-Alexandre and d'Ascoli, St{\'e}phane and Lample, Guillaume and Charton, Fran{\c{c}}ois},
  journal={Advances in Neural Information Processing Systems},
  volume={35},
  pages={10269--10281},
  year={2022}
}

@article{willard2022integrating,
  title={Integrating scientific knowledge with machine learning for engineering and environmental systems},
  author={Willard, Jared and Jia, Xiaowei and Xu, Shaoming and Steinbach, Michael and Kumar, Vipin},
  journal={ACM Computing Surveys},
  volume={55},
  number={4},
  pages={1--37},
  year={2022},
  publisher={ACM New York, NY}
}

@article{bergen2019machine,
  title={Machine learning for data-driven discovery in solid Earth geoscience},
  author={Bergen, Karianne J and Johnson, Paul A and de Hoop, Maarten V and Beroza, Gregory C},
  journal={Science},
  volume={363},
  number={6433},
  pages={eaau0323},
  year={2019},
  publisher={American Association for the Advancement of Science}
}

@book{hsieh2009machine,
  title={Machine learning methods in the environmental sciences: Neural networks and kernels},
  author={Hsieh, William W},
  year={2009},
  publisher={Cambridge university press}
}

@article{dumka2022implementation,
  title={Implementation of Buckingham's Pi theorem using Python},
  author={Dumka, Pankaj and Chauhan, Rishika and Singh, Ayush and Singh, Gaurav and Mishra, Dhananjay},
  journal={Advances in Engineering Software},
  volume={173},
  pages={103232},
  year={2022},
  publisher={Elsevier}
}

@article{ali2021unraveling,
  title={Unraveling the combined actions of a Holling type III predator--prey model incorporating Allee response and memory effects},
  author={Ali, Md Ramjan and Raut, Santanu and Sarkar, Susmita and Ghosh, Uttam},
  journal={Computational and Mathematical Methods},
  volume={3},
  number={2},
  pages={e1130},
  year={2021},
  publisher={Wiley Online Library}
}

@article{fan2021interpretability,
  title={On interpretability of artificial neural networks: A survey},
  author={Fan, Feng-Lei and Xiong, Jinjun and Li, Mengzhou and Wang, Ge},
  journal={IEEE Transactions on Radiation and Plasma Medical Sciences},
  volume={5},
  number={6},
  pages={741--760},
  year={2021},
  publisher={IEEE}
}

@article{brunton2016discovering,
  title={Discovering governing equations from data by sparse identification of nonlinear dynamical systems},
  author={Brunton, Steven L and Proctor, Joshua L and Kutz, J Nathan},
  journal={Proceedings of the national academy of sciences},
  volume={113},
  number={15},
  pages={3932--3937},
  year={2016},
  publisher={National Academy of Sciences}
}

@article{korns2013baseline,
  title={A baseline symbolic regression algorithm},
  author={Korns, Michael F},
  journal={Genetic Programming Theory and Practice X},
  pages={117--137},
  year={2013},
  publisher={Springer}
}

@article{sun2022symbolic,
  title={Symbolic physics learner: Discovering governing equations via monte carlo tree search},
  author={Sun, Fangzheng and Liu, Yang and Wang, Jian-Xun and Sun, Hao},
  journal={arXiv preprint arXiv:2205.13134},
  year={2022}
}

@article{buckingham1914physically,
  title={On physically similar systems; illustrations of the use of dimensional equations},
  author={Buckingham, Edgar},
  journal={Physical review},
  volume={4},
  number={4},
  pages={345},
  year={1914},
  publisher={APS}
}

@article{bakarji2022dimensionally,
  title={Dimensionally consistent learning with Buckingham Pi},
  author={Bakarji, Joseph and Callaham, Jared and Brunton, Steven L and Kutz, J Nathan},
  journal={Nature Computational Science},
  volume={2},
  number={12},
  pages={834--844},
  year={2022},
  publisher={Nature Publishing Group US New York}
}

@article{virtanen2020scipy,
  title={SciPy 1.0: fundamental algorithms for scientific computing in Python},
  author={Virtanen, Pauli and Gommers, Ralf and Oliphant, Travis E and Haberland, Matt and Reddy, Tyler and Cournapeau, David and Burovski, Evgeni and Peterson, Pearu and Weckesser, Warren and Bright, Jonathan and others},
  journal={Nature methods},
  volume={17},
  number={3},
  pages={261--272},
  year={2020},
  publisher={Nature Publishing Group US New York}
}

@article{qi2025ndawl,
  title={NDAWL-PINN: a new non-dimensionalization and multi-task learning approach for efficient training of physics-informed neural networks to solve the shallow water equations},
  author={Qi, Xin and Zhang, Dawei and Wang, Fan and Bi, Wuxia and Lu, Mingda},
  journal={Engineering Applications of Computational Fluid Mechanics},
  volume={19},
  number={1},
  pages={2535015},
  year={2025},
  publisher={Taylor \& Francis}
}

@article{yang2024data,
  title={Data-driven dryout prediction in helical-coiled once-through steam generator: A physics-informed approach leveraging the Buckingham Pi theorem},
  author={Yang, Kuang and Liao, Haifan and Xu, Bo and Chen, Qiuxiang and Hou, Zhenghui and Wang, Haijun},
  journal={Energy},
  volume={294},
  pages={130822},
  year={2024},
  publisher={Elsevier}
}

@article{chandra2024role,
  title={Role of physics in physics-informed machine learning},
  author={Chandra, Abhishek and Bakarji, Joseph and Tartakovsky, Daniel M},
  journal={Journal of Machine Learning for Modeling and Computing},
  volume={5},
  number={1},
  year={2024},
  publisher={Begel House Inc.}
}

@article{paszke2019pytorch,
  title={Pytorch: An imperative style, high-performance deep learning library},
  author={Paszke, Adam and Gross, Sam and Massa, Francisco and Lerer, Adam and Bradbury, James and Chanan, Gregory and Killeen, Trevor and Lin, Zeming and Gimelshein, Natalia and Antiga, Luca and others},
  journal={Advances in neural information processing systems},
  volume={32},
  year={2019}
}

@article{srbench,
  title={A unified framework for deep symbolic regression},
  author={Landajuela, Mikel and Lee, Chak Shing and Yang, Jiachen and Glatt, Ruben and Santiago, Claudio P and Aravena, Ignacio and Mundhenk, Terrell and Mulcahy, Garrett and Petersen, Brenden K},
  journal={Advances in Neural Information Processing Systems},
  volume={35},
  pages={33985--33998},
  year={2022}
}

@article{srbench++,
  title={SRBench++: principled benchmarking of symbolic regression with domain-expert interpretation},
  author={de Franca, Fabricio O and Virgolin, Marco and Kommenda, M and Majumder, MS and Cranmer, M and Espada, G and Ingelse, L and Fonseca, A and Landajuela, M and Petersen, B and others},
  journal={IEEE transactions on evolutionary computation},
  year={2024},
  publisher={IEEE}
}

@article{holling1959some,
  title={Some characteristics of simple types of predation and parasitism1},
  author={Holling, Crawford S},
  journal={The canadian entomologist},
  volume={91},
  number={7},
  pages={385--398},
  year={1959},
  publisher={Cambridge University Press}
}

\end{document}